\documentclass{article}

\usepackage{arxiv}

\usepackage[utf8]{inputenc} 
\usepackage[T1]{fontenc}    
\usepackage[colorlinks,allcolors=blue]{hyperref}
\usepackage{url}            
\usepackage{booktabs}       
\usepackage{amsfonts}       
\usepackage{nicefrac}       
\usepackage{microtype}      
\usepackage{graphicx}
\usepackage{natbib}
\usepackage{doi}
\usepackage{amssymb}
\usepackage{tabularx}
\usepackage{multirow}
\usepackage{multicol}
\usepackage{subfig}
\usepackage{caption}
\usepackage{tikz}
\usepackage{color}
\usepackage{xspace}
\usepackage{enumerate}
\usepackage{xurl}
\usepackage[toc]{appendix}

\title{Common Practices and Taxonomy in Deep Multi-view Fusion for Remote Sensing Applications}

\date{} 					

\author{ 
    Francisco Mena$^{1,2}$\thanks{\href{https://github.com/fmenat}{github.com/fmenat}}\href{https://orcid.org/0000-0002-5004-6571}{\includegraphics[scale=0.06]{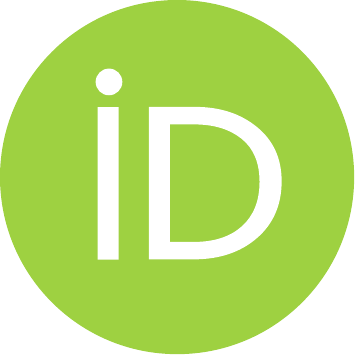}\hspace{1mm}},
	Diego Arenas$^{2}$\href{https://orcid.org/0000-0001-7829-6102}{\includegraphics[scale=0.06]{figs/orcid.pdf}\hspace{1mm}}, 
	Marlon Nuske$^{2}$\href{https://orcid.org/0000-0002-0651-0664}{\includegraphics[scale=0.06]{figs/orcid.pdf}\hspace{1mm}}, 
	Andreas Dengel$^{1,2}$\href{https://orcid.org/0000-0002-6100-8255}{\includegraphics[scale=0.06]{figs/orcid.pdf}\hspace{1mm}} \\
	$^1$Department of Computer Science,	Technical University of Kaiserslautern (TUK), Kaiserslautern, Germany;\\
	$^2$Smart Data and Knowledge Services, German Research Center for Artificial Intelligence (DFKI),\\ Kaiserslautern, Germany.\\
	\texttt{menatoro@rhrk.uni-kl.de}, \{\texttt{diego.arenas, marlon.nuske, andreas.dengel}\}\texttt{@dfki.de} \\
}

\renewcommand{\headeright}{A Preprint of a Review}
\renewcommand{\undertitle}{Preprint submitted to Journal}
\renewcommand{\shorttitle}{Deep Multi-view Fusion for Remote Sensing}

\hypersetup{
pdftitle={Deep Multi-view Fusion for Remote Sensing Applications},
pdfsubject={q-bio.NC, q-bio.QM},
pdfauthor={Francisco Mena, Diego Arenas, Marlon Nuske, Andreas Dengel},
pdfkeywords={Multi-view Learning, Multi-modal Learning, Data Fusion, Remote Sensing, Deep Learning, Supervised Learning}
}

\begin{document}
\maketitle

\begin{abstract}
The advances in remote sensing technologies have boosted applications for Earth observation. These technologies provide multiple observations or views with different levels of information.
They might contain static or temporary views with different levels of resolution, in addition to having different types and amounts of noise due to sensor calibration or deterioration.
A great variety of deep learning models have been applied to fuse the information from these multiple views, known as deep multi-view or multi-modal fusion learning. 
However, the approaches in the literature vary greatly since different terminology is used to refer to similar concepts or different illustrations are given to similar techniques.
This article gathers works on multi-view fusion for Earth observation by focusing on the common practices and approaches used in the literature. We summarize and structure insights from several different publications concentrating on unifying points and ideas.
In this manuscript, we provide a harmonized terminology while at the same time mentioning the various alternative terms that are used in literature.
The topics covered by the works reviewed focus on supervised learning with the use of neural network models.
We hope this review, with a long list of recent references, can support future research and lead to a unified advance in the area
\end{abstract}


\section{Introduction} \label{sec:intro}
Earth observation (EO) allows the study and analysis of different aspects of human life and natural resources, where remote sensing (RS) technologies are a crucial factor in providing a global perspective on the Earth. The final purpose is to make better data-informed decisions based on the current and future state of the planet. Many applications in this context express phenomena or objects that could be described or represented by multiple observations. For instance, crop-type that could be classified based on observations from different satellites \citep{ofori-ampofo2021croptypemappingb}, an agricultural yield that could be estimated based on ground-based weather and RS-based optical information \citep{meng2021predictingmaizeyielda}, or evapotranspiration (water evaporation into the atmosphere) that could be predicted based on different meteorological factors \citep{kumar2008comparativestudyconventional}. 
Deep learning, i.e. deep neural networks, allows learning complex functions as predictive models on possible non-linear and heterogeneous patterns. Therefore, they have been successfully applied to different areas of EO \citep{yuan2020deeplearningenvironmentala}. 
However, the main difference from a comprehensive learning scenario is that the object of interest could be represented by multiple observations, views, or modalities. This challenge makes it necessary to propose appropriate approaches or models to combine these various types of information.

There are many challenges when combining multiple views, e.g. the heterogeneity of the views. Since data in the EO domain has four types of sensor resolutions: spectral, spatial, temporal, and radiometric, it makes difficult to combine them in each use-case. Another reason corresponds to the source from which the multi-view (MV) data was obtained, causing them to have different structures and amounts of noise \citep{wang2020whatmakestraininga}. 
Besides, machine learning models are not perfect and make errors, e.g. from inductive biases (the errors due to the learning from empirical data). 
Therefore, if we try to include too many views, the model may collapse due to over-parameterization \citep{wang2020whatmakestraininga} or the curse of dimensionality \citep{ghamisi2019multisourcemultitemporaldata,kang2020comparativeassessmentenvironmentala}.
Then, the goal of the multi-view learning models is to extract the most information from the available views, being able to supplement some missing information. 
This manuscript mainly focuses on deep learning models addressing the challenges of model design. What are the options for modeling? Which types of architecture and strategies of fusion to use? What are the common approaches in literature? These are the type of questions that we are aiming to answer in this manuscript.

The document outline is organized as follows. In section \ref{sec:literature}, some preliminary concepts with the conceptual framework are introduced. Followed by section \ref{sec:challenges}, with the challenges of the MV learning scenario, and section \ref{sec:questions}, where different questions are commented on based on how they are approached in the literature. 
In section \ref{sec:addcomp}, common approaches (characterized by the components used) are highlighted.
A brief discussion is presented on the use of specific views in the MV learning scenario in section \ref{sec:indviews}. Finally, section \ref{sec:concl} summarizes some final remarks and conclusions about the reviewed literature.

\section{Literature Overview} \label{sec:literature}
The main purpose of data fusion in multi-view (MV) learning is to combine the information of different perspectives (views) to provide a broader understanding of the phenomena and improve the prediction \citep{li2019surveymultiviewrepresentation}. However, the goal could be just to get an embedding to search for similar views, as is the case of MV alignment or representation learning \citep{hotelling1936relationstwosets,heidler2021selfsupervisedaudiovisualrepresentation}. This alignment is related to the area of contrastive learning \citep{chen2020simpleframeworkcontrastive}, where a model for each view is used to project the data into a shared sub-space. Nevertheless, this article only covers data fusion within the MV learning topic.

As the EO domain is an intrinsically MV scenario for machine learning models, many fusion approaches have been proposed in the literature. These approaches are sometimes similar but with different appearances, either concepts or designs. 
For instance, Sentinel-1 and Sentinel-2 satellites are referred to as multi-sensor \citep{waske2007fusionsupportvector,chen2017deepfusionremotea}, multi-source \citep{ienco2019combiningsentinel1sentinel2b}, or as multi-modal \citep{cuelarosa2021investigatingfusionstrategiesa,hong2021morediversemeansa} for land-use and land-cover classification (aka LULC). Another case is the reference to fusion after extracting a representation for each view by using neural networks, such as middle-level fusion \citep{wu2021convolutionalneuralnetworks}, layer-level fusion \citep{ofori-ampofo2021croptypemappingb}, or late-level fusion \citep{saintefaregarnot2022multimodaltemporalattention}.
This manuscript tries to provide a unified taxonomy and common practices found in the literature, providing a discussion about the advantages, disadvantages, and limitations of the different practices.

Some works already provide valuables review of MV learning models. Such as \cite{sun2013surveymultiviewmachine}, which focused on unsupervised and semi-supervised MV learning with a perspective on the theoretical work behind it. Later \cite{zhao2017multiviewlearningoverview} provided a review on the same line with updated references and interesting open problems in the area. \cite{lahat2015multimodaldatafusion} provided a review focused on multi-sensor, medical, and environmental applications. 
More recently, with the advances of deep learning models, some studies review and compare conventional (without neural networks) with deep models in the MV learning topic \citep{ramachandram2017deepmultimodallearning,yan2021deepmultiviewlearning,summaira2021recentadvancestrends}. In these works several public MV datasets are shared which are mostly based on human and action recognition. 
On the other hand, there are surveys focused on the EO domain. For example, \cite{gomez-chova2015multimodalclassificationremotea} presented a review in the context of RS image classification, while in \cite{zhu2018spatiotemporalfusionmultisourcea,ghamisi2019multisourcemultitemporaldata} works they focused on RS image fusion. Image fusion in the sense of spatio-temporal fusion before any learning for a downstream task \citep{gao2006blendinglandsatmodisa}. \cite{salcedo-sanz2020machinelearninginformationa} reviewed several data fusion methods in different applications and use-case of the EO domain.
Recently, \cite{li2022deeplearningmultimodal} gathered open-source code and RS benchmark datasets for some EO applications in data fusion for multi-modal learning. In this review, they focused on several specific examples categorized into two main types of data fusion: homogeneous vs heterogeneous.
However, the proposed review is motivated by sharing and unifying the common practices and terminology, in addition to highlighting the most promising approaches for MV fusion learning in the EO domain.

The papers in this review were selected based on a query search with the following keywords: ``data fusion'', ``multi-view learning'', ``multi-modal learning'', ``multi-source'', ``multi-sensor'', ``earth observation'', ``remote sensing''. The journal sources from where the papers were obtained are: \textit{Sensors}
, \textit{Remote Sensing}
, \textit{Remote Sensing of Environment}
, \textit{International Journal of Applied Earth Observation and Geoinformation}
, \textit{IEEE Transactions on Geoscience and Remote Sensing}
, \textit{IEEE Journal of Selected Topics in Applied Earth Observations and Remote Sensing}
, \textit{IEEE Geoscience and Remote Sensing Letters}
, \textit{Environmental Research}
, \textit{International Journal of Remote Sensing}
. While the paper's archive used to perform the search are: Google Scholar\footnote{\url{https://scholar.google.cl/}} and Astrophysics Data System (ADS)\footnote{\url{https://ui.adsabs.harvard.edu}}, obtaining 100 papers in MV learning for EO that are reviewed in this manuscript.

\subsection{Background Concepts} \label{sec:background}

\textbf{Earth observation} (EO) is the gathering and study of information about the biological, chemical, and physical systems of the planet Earth. It can be performed via remote-sensing (RS) technologies and ground-based (in-situ) techniques or stations. \textbf{Remote sensing} usually involves observing objects from platforms that are distant from the object being observed, e.g. satellites, aerial images, or unmanned aerial vehicles (UAV). This technology allows the collection of data from sensors with different types of resolution: Spectral, as electromagnetic bands or channels obtained by different filters; Spatial, as the area covered by each pixel based on the distance of the object to the observed area or the resolution of the instrument; Temporal, as the frequency with which an area is swept, and Radiometric, affected by the sensor sensibility, calibration, and deterioration.
 
There are plenty of EO applications and tasks where machine learning models have been applied. The most common tasks explored in the literature are regression, classification, and segmentation. 
In regression, the task is to predict a continuous value from EO data, e.g. estimate the crop yield produced in a particular field during a growing season (agricultural yield prediction), estimate the precipitation for future days (precipitation forecasting), or estimate the snow depth of a region.
In classification, the task is to predict a class within a set of possible labels for a particular region, e.g. identify a potential target in the Earth's surface (automatic target recognition), identify a plant or crop growing in a field (vegetation recognition), or identify if a field is irrigated or not (irrigated recognition).
In segmentation, the task is to predict a class in a mesh of a particular region (as a pixel-wise classification), e.g. identify the type of use of a region (e.g. LULC mapping), identify which section of a region is flooded (flood mapping), or identify which pixels of an RS image are covered by clouds (cloud segmentation).

An individual \textbf{view} is a data, observation, information channel, or feature set associated with an object of interest that contains some information about it (direct or indirect) \citep{hotelling1936relationstwosets, blum1998combininglabeledunlabeled}.
In EO, the most common views come from passive observations, which measure the solar radiation reflection on the Earth's surface, also referred to as optical images. The most common options for the optical view are: RGB (red-green-blue bands), multi-spectral (MS, with more bands on the spectrum than RGB), hyper-spectral (HS, with more than hundreds of bands), or panchromatic (PAN, a single band with a broad spectrum). Sentinel-2, Landsat-8, MODIS, or custom UAV are well-known sources to obtain an optical view. There are several examples of spatial indexes calculated from the optical views \citep{kim2016machinelearningapproaches,johnson2016cropyieldforecasting,gomez2019potatoyieldpredictionb,konapala2021exploringsentinel1sentinel2}, from which the most common for agricultural purposes is the NDVI (normalized difference vegetation index) and EVI (enhanced vegetation index).
Other types of views are based on active observation, which involve sending electromagnetic pulses to the Earth's surface and recording the reflected energy. Two commonly used views on this are synthetic aperture radar (SAR) which uses radio or microwave, and light detection and ranging (LiDAR), which uses infrared waves to emit pulses. Sentinel-1 or MODIS are example sources of the SAR view, while private UAV can be used to obtain a LiDAR view.
From this active observation, it is possible to generate a digital elevation map/model (DEM) that represents the surface topography of the Earth. There are two types of DEM: digital surface models (DSMs) and digital terrain models (DTMs). DSMs contain only the bare ground, while DSM additionally contains objects such as vegetation and buildings. In EO, DSMs are the commonly used view.
Lastly, some meteorological variables represent one or multiple views (depending on each case) and could be obtained from RS or ground-based instruments. For example, temperature, precipitation, solar radiation, wind speed, humidity, and vapor pressure.
 
\textbf{Multi-view} (MV) learning considers the learning scenario where multiple feature sets (views) of the same object are available, usually under the assumption that all the views are available for each object. There are different types of MV data in EO as the examples stated above.  
In this publication, we use the ``multi-view'' term as a general concept that summarizes all different concepts used in literature: multi-modal, multi-source, or multi-sensor. 
This concept does not constrain that views have to be complementary, represent different physical quantities \citep{lahat2015multimodaldatafusion}, or be from different data-type (e.g. images, signals, metadata). 
For instance, some MV data can be different multi-band images \citep{ienco2019combiningsentinel1sentinel2b}, different groups of spectral bands, such as RGB and NIR (Near Infra-Red) \citep{sa2016deepfruitsfruitdetection}, or additional views extracted from the optical view, such as NDVI \citep{audebert2017semanticsegmentationeartha,audebert2018rgbveryhigh}, or neural network features \citep{nijhawan2017deeplearninghybrid}.

In this article, we focus on \textbf{data fusion} within the context of MV learning. 
In literature, the term data fusion is commonly used in the context of database management, referring to integrating multiple heterogeneous data sources into a single, consistent and clean representation \citep{bleiholder2009datafusion}. 
In EO, image fusion usually refers to combining the geometric detail of multi-band images to produce a final image that enhances the good properties of these before learning any downstream task \citep{zhu2018spatiotemporalfusionmultisourcea,camps-valls2021deeplearningearth}, e.g. spatio-temporal fusion \citep{gao2006blendinglandsatmodisa,scarpa2018cnnbasedfusionmethod,lei2022convolutionneuralnetwork} or spatio-temporal-spectral fusion \citep{shen2016integratedframeworkspatioa}. 
However, in MV learning, data fusion could be a diffuse concept that takes different interpretations depending on the application. Therefore, our emphasis is on the following interpretation: integrate and merge the information of MV data with machine learning models to maximize the predictive performance on a given downstream task.
There could be other goals studied in MV learning for EO that are not covered within this manuscript.
The MV alignment or representation learning consists of learning a shared sub-space where the views could be projected, similar to contrastive learning \citep{chen2020simpleframeworkcontrastive}. What is explored is how to project the views to obtain low-dimensional vectors (with continuous or discrete variables) that could be used later, e.g. for cross-modal retrieval \citep{chen2020deepcrossmodalimage,cheng2021deepsemanticalignment}.

\textbf{Deep learning} refers to the use of neural networks (NNs) as machine learning models, usually fed with raw-format data, such as images or time series.
These models could be trained in a supervised (with labels) or unsupervised (without labels) way. When training in a supervised fashion, it is usually for a \textbf{downstream task} (aka predictive task or learning task), which is some task that the model needs to learn to predict through minimizing a \textbf{predictive loss}. Thus, a \textbf{predictive model} (aka prediction model or classifier) is used to give the final decision or prediction for the task based on the input data. 
However, when the models learn from raw data, usually an \textbf{encoder} model (aka backbone or extraction network) is used. These encoder models obtain a representation that compresses the most valuable information from the input, which feeds the predictive model.
The standard NN architecture used as a predictive model is the multi-layer perceptron (MLP), which incorporate fully connected layers. While the standard network architecture used as an encoder of image or spatial information is a convolutional neural network (CNN). But, when learning on sequential data, recurrent neural networks (RNN) with LSTM or GRU layers are usually used. 
We refer to predictive performance as the predictive accuracy of the supervised trained model.
When training in an unsupervised fashion, it is usually to extract useful representations of the data, which could be used for analysis or the downstream task, e.g. by self-supervision \citep{chen2020simpleframeworkcontrastive}.

\subsection{History of Multi-view Learning}
MV learning is a research area that began several years ago with the mathematical statistician \textit{Hotelling} and the CCA (correlation canonical analysis) proposal in 1936 \citep{hotelling1936relationstwosets}. In this work, linear mapping is learned to maximize the correlation between two views, on which the research continued.
For instance, \cite{kettenring1971canonicalanalysisseveral} extended CCA to multiple feature sets in 1971. 
More recently, \cite{andrew2013deepcanonicalcorrelationa} used NN models to learn non-linear mappings and correlations in 2013. 
However, up to our knowledge, the work of \cite{blum1998combininglabeledunlabeled} in 1998 is the first to mentioned the ``view'' concept and provided the first theoretical foundations of MV learning in classification, which was extended by \cite{muslea2002activesemisupervisedlearning}.
Furthermore, the first mention that we found to ``multi-view'' in the EO domain is associated with LULC application in 2015 \citep{luus2015multiviewdeeplearning}, where different scales (zooms) of an image were used as MV data. 
The ``multi-view'' concept was also used in the recent area of contrastive learning, having applications in the EO domain \citep{stojnic2021selfsupervisedlearningremote,heidler2021selfsupervisedaudiovisualrepresentation}.

Deep MV learning has been deeply explored in EO thanks to advances in NN models and open science culture \citep{camps-valls2021deeplearningearth}. EO domain is usually an open big data source that has been wrapped in a community of generating open access code, benchmark datasets, or pre-trained models. 
As in other areas of machine learning, the research started by using machine learning models that learn from tabular data (e.g. metadata). To give an idea, linear models such as perceptron in 1989 \citep{benediktsson1989neuralnetworkapproaches} and SVM (support vector machine) in 2006-2007 \citep{nemmour2006multiplesupportvector,waske2007fusionsupportvector} for LULC, or non-linear models such as decision trees in 2006 \citep{gomez-chova2006urbanmonitoringusing} and MLP in 2008 \citep{russ2008dataminingneural}. Later the community began to explore highly non-linear functions through deep NNs. For example, using multiple CNN models for each view in LULC during 2016-2017 \citep{marmanis2016semanticsegmentationaerial,chen2017deepfusionremotea,audebert2017semanticsegmentationeartha}.
In the book \citep{camps-valls2021deeplearningearth} it has been noted three different phases of research in the EO domain when using NNs. Exploration (from 2014), where the first NN models began to apply to different EO tasks. Benchmarking (from 2016), where standard datasets were released and started used as sources to validate and compare. Lastly, EO-driven methodological research (from 2019), where proposals focused beyond the applicability towards other aspects such as uncertainty or reasoning.

\section{Challenges of Deep Multi-view Learning} \label{sec:challenges}
In this section, we will discuss some intrinsic challenges of MV learning in the EO domain from the perspective of deep MV learning models. For a comprehensive review of the deep MV learning topic (NN models in the MV learning), please refer to \cite{ramachandram2017deepmultimodallearning}.

\cite{wang2020whatmakestraininga} presented interesting insights into the difficulties to train a MV learning model focused on the vision domain. The results also apply to the EO domain, which we comment on below.
\begin{enumerate}
    \item Heterogeneity of models: Since views could have different resolutions (e.g. spatial and temporal), each one will require different network architectures to process them. To give an idea, a CNN for spatial data, RNN for temporal data, or MLP for tabular data in the same MV learning model. 
    \item Different information levels: Since views could have information at different levels (e.g. high-level vs low-level feature, or high vs low noise), each one will require different network complexity to process them. For example, an optical image might require more layers in the NN than a cloud mask. 
    \item More patterns: By feeding more views as input (patterns) to the network, the model needs to determine the most efficient way to relate these views to the output. For instance, the well-known curse of dimensionality\footnote{In machine learning, the curse of dimensionality can manifest itself through the decrease in predictive performance when increasing the number of input-features (dimensionality).} \citep{ghamisi2019multisourcemultitemporaldata} is something that could occur when RS views are included and concatenated \citep{kang2020comparativeassessmentenvironmentala}.
    \item Over-fitting: When having multiple networks (one for each view), the number of parameters gets increased with respect to having a single network (single-view learning). This scenario could cause over-fitting if there is not enough labeled data to learn from. 
\end{enumerate}
In addition, learning from empirical data can cause the models to fail on future test data. Therefore, the amount of labeled data for supervised training is crucial to reduce errors when the model is deployed. Besides, MV learning models tend to over-fit more than single-view models \citep{wang2020whatmakestraininga}, not only due to the number of parameters but also because each view over-fits and generalizes at different levels. However, MV models usually learn in a single optimization framework that does not consider these differences.
However, there could be shared (or common) and private (or complementary) information in each view \citep{blum1998combininglabeledunlabeled, castanedo2013reviewdatafusion} that is usually ignored in the design and included in the model bias.
\cite{christoudias2012multiviewlearningpresence} explored the problem when the complementary information between the views is high enough to have view disagreement. View disagreement consists that views expressing contradictory information about the ground truth \citep{christoudias2012multiviewlearningpresence}, causing problems in training. Ideally, one looks for views with a good balance between complementary and similar/correlated information.
Furthermore, determining the optimal number of views needed to describe an object under study can be difficult on the context of machine learning models, How can we be sure that by giving it more views the model will not get better or worse?
Everything mentioned above suggests that MV learning is not just having multiple models as a multi-branch network, but involves more things that have been approached in the literature.

\section{Main questions} \label{sec:questions}
As mentioned in the previous section, learning from multiple views is a complex task for learning models which might require important decisions by the \textit{practitioner}. We refer to the ``practitioner'' as the person in charge of the model design and experimentation in the machine learning context.
Many works have shown that the fusion choices affect predictive performance, such as where to fuse, which function to use, or include components (e.g. prediction correction, losses, or cross-modules). In the following, we will discuss some relevant points on this topic. 

\subsection{Where to Fuse?}

\begin{figure}[t]
\includegraphics[width=0.95\textwidth, page=1]{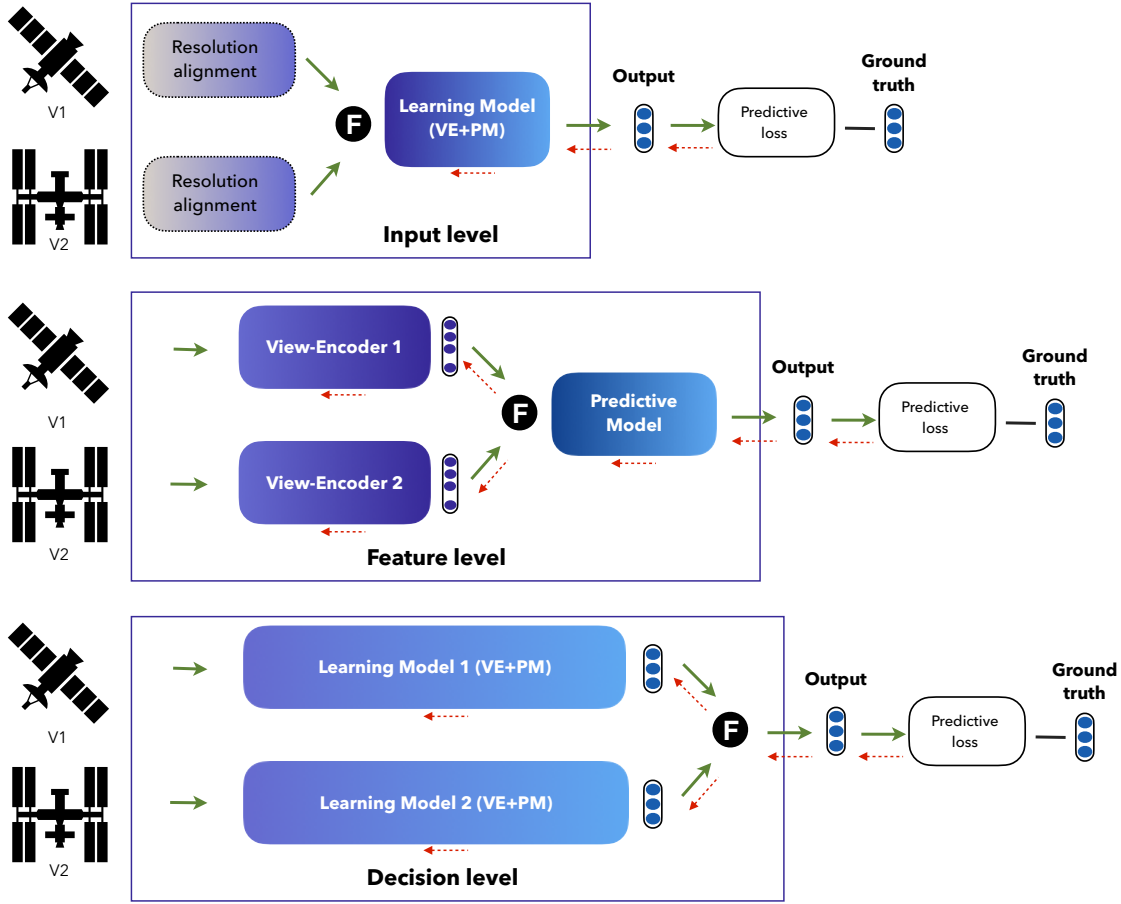}
\caption{Illustration of three different fusion strategies: input-level fusion at the top, feature-level fusion at the middle, and decision-level fusion at the bottom. The forward pass of the model is from left to right (green arrows), while the backward pass is from right to left (red dashed arrows). VE stands for view-encoder and PM for predictive model.}\label{fig:fusion_types}
\end{figure}

When using MV data for a downstream task, the question emerges at which stage of the deep learning model to fuse, with options such as early, middle, or late stages. For instance, some works \citep{zhang2020combiningopticalfluorescencea,ofori-ampofo2021croptypemappingb,hong2021morediversemeansa,saintefaregarnot2022multimodaltemporalattention} suggest fusion at: the first layers of the network (early), the middle layers (hidden features) of the network, or close to the decision of the network (late). However, early/middle/late concepts could be ambiguous in the context of NNs. \textit{Which layers define the boundaries into the partition?} as this is a hard question that will depend on the application and use-case, we present a more specific division inspired by the literature, see Figure~\ref{fig:fusion_types} for a visual example.
We categorized (not exclusively) the 100 papers reviewed for this division (see Table~\ref{sup:tab:fusion_strategies}).

\unskip

\textbf{Input-level} fusion (aka early-level and data-level fusion), is a naive approach that involves a direct concatenation of the data and feeding into a single model, i.e. it uses a single-view learning model. Before the concatenation, a resolution alignment step is usually required that matches all the dimensions of the views (see Figure~\ref{fig:fusion_types} for an illustration), e.g. spatio-temporal alignment using re-sampling or interpolation operations, or even more sophisticated operations like feature extraction. 
This is the most common strategy in different EO applications, with 49 papers.

\textbf{Feature-level} fusion (aka middle-level, intermediate-level, and layer-level fusion), focuses on obtaining a joint representation (ideally compact) that is useful for a prediction task. It considers NN view-encoders (or back-bone models) that generate a new representation for each view followed by a fusion or merge module and a predictive model  (see Figure~\ref{fig:fusion_types} for an example). This strategy was identified in 36 papers.

\textbf{Decision-level} fusion (aka late-level and classifier fusion) focuses on combining view-based predictions (e.g. probabilities, logits, or numerical values) obtained by parallel single-view learning models that process each view and yield a decision (see Figure~\ref{fig:fusion_types} for an example). This is the least used strategy in the literature reviewed, with 7 papers.
As shown in Figure~\ref{fig:fusion_types}, the ones that use multiple models to process each view are feature and decision-level fusion, while input-level fusion uses only a single model.
 
Moreover, the feature-level fusion might be further divided into two fusion sub-categories based on the types of layers used in the NNs. Fusion of sub-features consists of fusing low-level features with temporal, spatial, or spectral dimensions, such as images or time-series features, which may contain high dimensionality. For example, \cite{audebert2018rgbveryhigh} fused feature maps (spatial features) inside convolutional blocks of CNNs.
The other case (and more common) is the fusion of embeddings, this involve fusing high-level features such as vectors, which may contain low dimensionality. For example, \cite{chen2017deepfusionremotea} fused vector features after extracting an embedding with CNN.

There is an additional approach used by some works \citep{waske2007fusionsupportvector,luus2015multiviewdeeplearning,ahmad2017cnnganbased,liu2018deepconvolutionalneural,valero2019sentinelclassifierfusion,robinson2021globallandcovermapping,ma2021outcome2021ieee,rashkovetsky2021wildfiredetectionmultisensor,li2022outcome2021ieee} that is referred to as late-fusion but is different from the one mentioned previously. We name the approach ``ensemble-based aggregation'' since it is based on a two step process. The first step is to train a model for each view (without fusion) independently, while the second step is on test time. After the models are trained, a prediction aggregation is defined that merges the view-based predictions (e.g. through the average or majority vote), similar to an ensemble framework. This case corresponds to a model-agnostic fusion, where the information of multiple views does not interact, nor the relationship between these is exploited, i.e. the fusion is detached from the learning.

\begin{figure}[t!]
\includegraphics[width=0.95\textwidth, page=3]{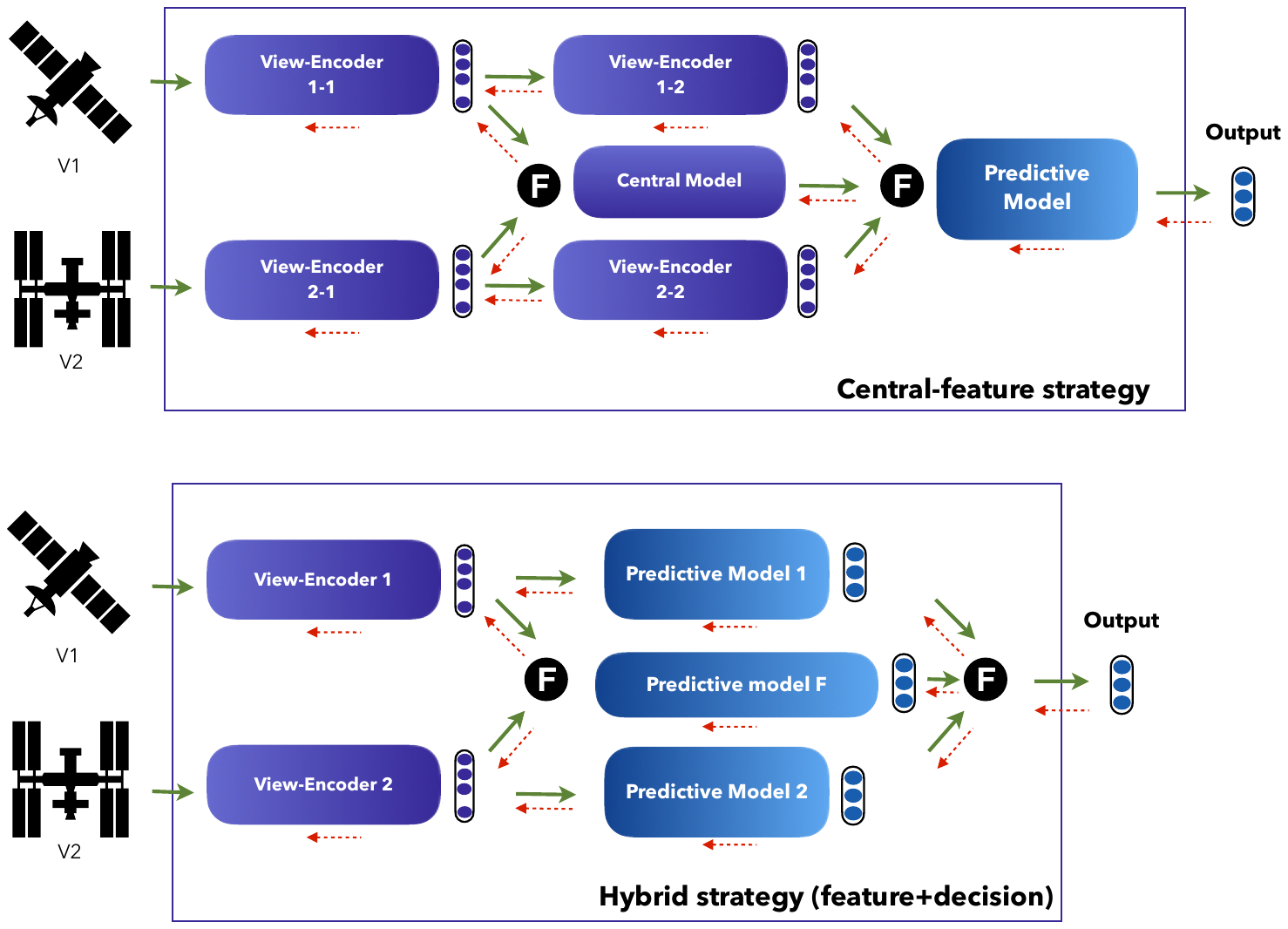}
\vspace{-1cm}\caption{Illustration of additional fusion strategies found in the literature: central-feature strategy at the top, and hybrid strategy at the bottom. The forward pass of the model is from left to right (green arrows), while the backward pass is from right to left (red dashed arrows). The ``predictive model F'' represent the predictive model that is fed with the fused representation.}\label{fig:fusion_types:extra}
\end{figure}   
\unskip
In addition, taking advantage of the flexibility of NN models, some works explore different fusion strategies. One case is hybrid fusion, which fuses the views at different stages of the NN. To illustrate, feature and decision fusions in the same model improve predictive performance with respect to only one fusion with passive and active based views  \citep{hang2020classificationhyperspectrallidar,zhang2020hybridattentionawarefusion,cuelarosa2021investigatingfusionstrategiesa}, or input and feature fusions in the same model \citep{cao2021integratingmultisourcedata,li2022outcome2021ieee}. However, it can also be fused in a dense-basis \citep{vielzeuf2018centralnetmultilayerapproach}, which means fusion across all layers of NN (usually with an additional central model that handles it), with some applications in EO \citep{pei2018sarautomatictarget,audebert2018rgbveryhigh,pei2018sarautomatictarget,zhang2020hybridattentionawarefusion,cao2021c3netcrossmodalfeaturea,zhou2022cegfnetcommonextraction,hosseinpour2022cmgfnetdeepcrossmodala,zhao2022multisourcecollaborativeenhanced}. An illustration of these approaches is shown in Figure~\ref{fig:fusion_types:extra}.

\begin{figure}[t!]
\includegraphics[width=0.65\textwidth, page=1]{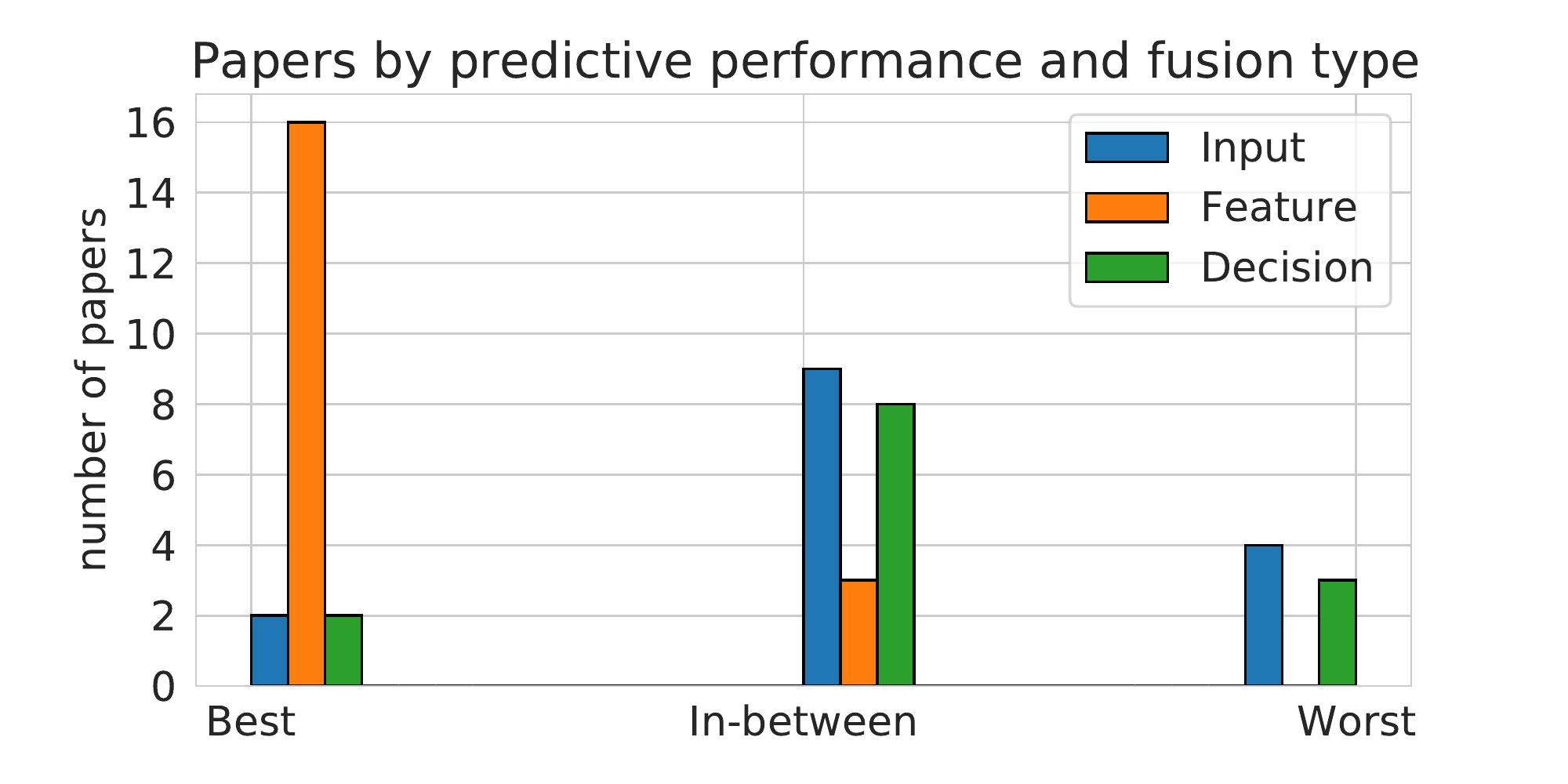}
\caption{The number of papers that show empirical evidence of being best, in-between or worst predictive performance within three fusion strategies (input, feature, and decision fusion). Individual papers are in Table~\ref{sup:tab:where_fuse}.}\label{fig:where_fuse}
\end{figure}
\unskip
\cite{hong2021morediversemeansa} also explored the question of where to fuse in NNs for the EO domain. However, conclusions were based on results of two EO datasets, while in this paper, a collection of different experimental evidence is gathered and summarized for comparison. 
In Figure~\ref{fig:where_fuse}, empirical evidence from the literature on different EO datasets (and tasks) is presented, comparing the three main fusion strategies discussed above (Figure~\ref{fig:fusion_types}). In this case, the comparison is categorical and corresponds to whether the fusion strategy shows the best, worst, or in-between predictive performance for the EO task.
The main outcome is that feature-level fusion has the best predictive performance in most of the cases compared with input and decision fusion \citep{chen2017deepfusionremotea,xu2018multisourceremotesensing,audebert2018rgbveryhigh,livieris2020multipleinputneuralnetworka,zhang2020hybridattentionawarefusion,maimaitijiang2020soybeanyieldpredictiona,srivastava2020finegrainedlandusecharacterization,wu2021convolutionalneuralnetworks,hong2021morediversemeansa,sebastianelli2021paradigmselectiondata,hosseinpour2022cmgfnetdeepcrossmodala,saintefaregarnot2022multimodaltemporalattention}, besides not having the worst predictive performance (at least in the evidence reviewed). Input-level and decision-level fusion strategies are quite competitive with each other, in some cases, input fusion is the best \citep{ofori-ampofo2021croptypemappingb,saintefaregarnot2022multimodaltemporalattention} (or worst \citep{wu2021convolutionalneuralnetworks,hong2021morediversemeansa,sebastianelli2021paradigmselectiondata}), and in others, decision fusion is the best \citep{sa2016deepfruitsfruitdetection,cuelarosa2021investigatingfusionstrategiesa} (or worst \citep{ofori-ampofo2021croptypemappingb,saintefaregarnot2022multimodaltemporalattention}), showing that the results strongly depend on the data. 
This evidence suggests that input and decision-level fusion are more uncertain techniques in terms of knowing a-priori if they will work better or worse in some EO applications. \cite{srivastava2020finegrainedlandusecharacterization} even compared empirically that when no fusion is performed, feature-based aggregation (sum of NN features previous to training) is better than ensemble-based aggregation (majority voting of predictions after training) when using RS-based and ground-based views for LULC.

In the following, some relevant points of the three main fusion strategies are commented on with respect to the modeling choices required by the practitioner:
\begin{enumerate}
    \item Input-level fusion: It has fewer model choices than the other fusion strategies (e.g. only one model), and it is simple and easily implemented. However, it requires defining the alignment step, which could add noise synthetically and requires domain knowledge. Besides being restricted to views from the same data type and cannot be initialized with pre-trained models (e.g. Imagenet transfer learning). 
    \begin{itemize}
        \item Advantages: Usually fewer parameters than in other fusion strategies since only one model is used. Besides, a complex model could (eventually) learn the relationship between views.
        \item Disadvantages: Fusion is not explicitly learned, high dimension of stacked data, the need to search for a complex model to exploit the relationships between views, and rely on the conditional independence of the source of information (however, this might not meet \citep{ramachandram2017deepmultimodallearning}).
    \end{itemize}
    \item Feature-level fusion: It avoids the alignment step but increases the model choices with respect to the input-level fusion (e.g. one choice for each view-encoder model). Besides, it needs to define the merge/fusion function.
    \begin{itemize}
        \item Advantages: Fusion is explicitly learned, enable to learn non-linear relationships and cross-view features between views (due to the information exchange), and allows to use of different model complexity (extraction ability) for each view.
        \item Disadvantages: Increase the number of parameters of the model.
    \end{itemize}
    \item Decision-level fusion: It also avoids the alignment step and increases the model choices with respect to input-level fusion.
    \begin{itemize}
        \item Advantages: Fusion is explicitly learned and is feature-independent, allows a non-linear mapping of the views, and uses different model complexity (extraction ability) for each view.
        \item Disadvantages: Increase (heavily) the number of parameters of the model, besides not being able to learn cross-modal features (due to the lack of feature exchange modules).
    \end{itemize}
\end{enumerate}

Since MV models tend to over-fit more than single-view models \citep{ienco2019combiningsentinel1sentinel2b,hong2021morediversemeansa}, the practitioner may find it difficult to choose models. However, \cite{sahu2021adaptivefusiontechniques} saw this point as an advantage, as the practitioner can assign a simple model for the view-encoders to avoid spending so many resources on that part and focus more on the fusion. 
Besides, the variations and improvements in predictive performance between the different fusion strategies (input/feature/decision) are quite slight compared to the time and resources spent on the search and training, while variations in data, such as including additional views, are sometimes more critical for improvement.

\subsection{How to Fuse?}
The practitioner must also define how the fusion or merge will be performed. As data fusion in deep MV learning involves NNs with all its design flexibility, many architectures with different ways to merge have been explored in the literature. For instance, use different types of merge functions such as uniform-sum, weighted-sum, product, concatenation, or with gated modules (as in adaptive fusion) \citep{arevalo2017gatedmultimodalunits,arevalo2020gatedmultimodalnetworksa}. Another case corresponds to the use of additional components or layers that perform the merge inside the model. This could be a central-model (see Figure~\ref{fig:fusion_types:extra}) \citep{audebert2018rgbveryhigh,zhao2022multisourcecollaborativeenhanced}, a model with cross-channels between the view-encoders \citep{mohla2020fusatnetdualattention,cao2021c3netcrossmodalfeaturea,hong2021morediversemeansa,hosseinpour2022cmgfnetdeepcrossmodala}, a model with average correction in decision-level fusion \citep{audebert2017semanticsegmentationeartha}, a RNN model for sequential views \citep{wang2021multiviewattentioncnnlstm}, or other cases. 
Since for input-level fusion, the straightforward merge function is concatenation, we will discuss in the following some common merge functions by focusing on the other cases.
Considering the representation (or prediction) obtained by $V$ view-encoders (or view-predictive models) on each view $\{z_v\}_{v=1}^V$ and the merge function $F(\cdot)$, the fused representation could be expressed by $z^* = F(\{z_1, z_2, \ldots, z_V\})$. Table~\ref{tab:merge_func} summarized some options for the function $F$ used in the literature, which allows a mathematical comparison. 

\begin{table}[!ht] 
\centering
\caption{Common merge functions used in the literature. With $z^{*} = F(\{z_1, z_2, \ldots, z_V\})$, $\odot$ as the Hadamard product (element-wise product), and $z^{'}$ an auxiliary joint representation obtained by some previous merge function. \textit{Num} corresponds to the number of papers that use that merge function, see Table~\ref{sup:merge_func} for the complete references.}\label{tab:merge_func}
\newcolumntype{L}{>{\raggedright\arraybackslash}X}
\begin{tabularx}{\textwidth}{llLll} \hline
\textbf{Name}   & \textbf{Function ($F$)}   & \textbf{Additional}  & \textbf{Mode} & \textbf{Num} \\ \hline
Concatenation & $\left[ z_1, z_2, \ldots, z_V \right]$ &  - & stack  & 36\\
Attention & $ \alpha \cdot \left[ z_1, z_2, \ldots, z_V \right]$ & $\alpha = G(z^{'}) \in [0,1]$  & stack & 5 \\
Uniform-sum & $\sum_{v} z_v$            & (optional) $z^{*}  = z^{*} / V$ & pool & 16 \\
Weighted-sum & $ \sum_v g_v \odot z_v$ & with $\sum_v g_v =1$ some learnable or fixed weights & pool & 7 \\ 
Gated & $\sum_v g_v \odot z_v $ & $g_v = G(z^{'})$, $\sum_v g_v =1$ & pool & 3 \\
Product & $\odot_{v} z_v$           & (optional) $z^{*}  = z^{*}/C$, with normalization factor $C$ & pool & 2\\
Maximum & $\max\{z_1, z_2, \ldots, z_V\}$ & - & pool  & 2 \\
Majority & $\mathrm{mode} \{z_1, z_2, \ldots, z_V\} $ & - & pool & 3 \\ \hline
\end{tabularx}
\end{table}
\unskip

In general, there is not much experimental evidence focused on merge function comparison, especially in the context of EO. 
While most of the merge functions produce a pooling aggregation (fused dimension is the same as individual-view dimensions), the most commonly used in literature, concatenation, performs a stack of the information with the joint dimension equals the sum of individual dimensions. However, a few works have shown that the pooling functions perform better than concatenation when fusing an optical-MS with a LiDAR view \citep{hang2020classificationhyperspectrallidar}, or an optical-RGB with a DEM view \citep{zhang2020hybridattentionawarefusion}.
When comparing different pooling aggregations, other publications show that the convex combination of a uniform-sum was better than the maximum for land-use characterization with RS and ground-based views \citep{srivastava2020finegrainedlandusecharacterization}, or than the product for crop-type mapping with optical-MS and SAR views \citep{ofori-ampofo2021croptypemappingb}, or than majority voting for building extraction with optical and DEM views \citep{zhang2020hybridattentionawarefusion}.
This could be due to the fact that convex combination is not such a strong pooling as maximum or product functions. This result is also evidenced in some works outside EO for object recognition \citep{feichtenhofer2016convolutionaltwostreamnetworka, vielzeuf2018centralnetmultilayerapproach,arevalo2020gatedmultimodalnetworksa}.
In addition, \cite{hong2021multimodalremotesensing} showed that view-specific features are more useful for prediction than view-shared features (through a shared model), suggesting a step towards exploiting the individual information within each view.
On the other hand, empirical evidence in different EO tasks shows that having two (hybrid fusion) \citep{audebert2018rgbveryhigh,cuelarosa2021investigatingfusionstrategiesa} or multiple (dense) \citep{hazirbas2017fusenetincorporatingdeptha,pei2018sarautomatictarget,zhang2020hybridattentionawarefusion} fusion channels improve over only one channel. These works give credit to the idea that the model has the ability to exchange more information from the views and correct the fusion of the earlier stages (this is just an assumption as it has not yet been proved in the literature).

\subsection{What to Focus?}
As a first concern, the practitioner could be curious about what part of the modeling to focus on and give more attention to. For example, in which part design a more complex architecture to obtain better predictive performance. Since views in EO data have different resolutions, e.g. an image, a sequence of images, a sequence of vectors, or metadata, it is necessary to define an appropriate view-encoder model to process each view.
The most common choice for NN-based view-encoders in EO is MLP or well-known models and architectures for the specific data-type \citep{nguyen2019spatialtemporalmultitasklearninga,maimaitijiang2020soybeanyieldpredictiona,livieris2020multipleinputneuralnetworka,gadiraju2020multimodaldeeplearninga,hong2021morediversemeansa}. For instance, when using optical images (usually for LULC), the common choice is to use known CNN architectures \citep{scott2018enhancedfusiondeep}, such as AlexNet \citep{valente2019detectingrumexobtusifolius,yalcin2018phenologyrecognitionusing,reyes2015finetuningdeepconvolutional,lee2015deepplantplantidentification,mohanty2016usingdeeplearning}, VGG \citep{chew2020deepneuralnetworksa,nowakowski2021croptypemappinga,bargoti2017deepfruitdetection,jiang2019novelcropweed}, InceptionNet/GoogleNet \citep{nowakowski2021croptypemappinga,mohanty2016usingdeeplearning,ramcharan2017deeplearningimagebased}, ResNet \citep{rudner2019multi3netsegmentingflooded,rambour2020flooddetectiontime,irvin2020forestnetclassifyingdrivers,ma2021outcome2021ieee,yuan2021deeplearningbasedmultispectralsatellite,zhou2022cegfnetcommonextraction}, DenseNet \citep{irvin2020forestnetclassifyingdrivers}, SENet \citep{ma2021outcome2021ieee}, and EfficienNet \citep{irvin2020forestnetclassifyingdrivers,dimartino2021multibranchdeeplearning}. While SegNet \citep{bosilj2020transferlearningcropa} and Faster R-CNN \citep{sa2016deepfruitsfruitdetection} have been used for optical-MS image segmentation. On the other hand, when using temporal data, RNN models (with LSTM or GRU) are usually selected \citep{gadiraju2020multimodaldeeplearninga,livieris2020multipleinputneuralnetworka}.

As occurs in other applications, the increase of complexity in MV models, e.g. layers and parameters (with the appropriate regularization techniques) commonly improves the predictive performance in EO tasks. Some examples are by increasing network layers and branches on LULC \citep{audebert2018rgbveryhigh}, or by increasing network parameters on automatic target recognition \citep{wang2021multiviewattentioncnnlstm}. However, some cases evidence the opposite, that having a less complex model achieve good predictive performance for MV learning in EO, e.g. by having fewer layers on LULC \citep{gadiraju2020multimodaldeeplearninga,hong2021morediversemeansa}.

In some works \citep{feichtenhofer2016convolutionaltwostreamnetworka,perez-rua2019mfasmultimodalfusion} we found the recommendation to pre-train the view-encoder or predictive models for each view independently in order to have them minded for the fusion in the downstream task. 
This use of pre-trained models for EO tasks (aka transfer learning) varies from fine-tuned models pre-trained on large image datasets outside the EO domain \citep{marmanis2016semanticsegmentationaerial,nijhawan2017deeplearninghybrid,srivastava2019understandingurbanlanduse}, to models pre-trained with the same task in a different geographical region \citep{wang2018deeptransferlearninga}, or models pre-trained on a different task in the same geographical region. For example, \cite{khaki2020cnnrnnframeworkcropa} proposed a crop-type transfer on the same region, i.e. fine-tuning in a different crop-type.

\cite{sahu2021adaptivefusiontechniques} mentioned that simple view-encoders could be combined with a complex fusion mechanism which can make it compete against complex single-view models (such as transformers or deep networks).
However, the empirical evidence of \cite{gadiraju2020multimodaldeeplearninga} for crop classification showed that having a linear model (SVM) after the fusion worked better than non-linear (MLP) by having complex view-encoders that process the views (optical-RGB image and NDVI time series).
\cite{ienco2019combiningsentinel1sentinel2b} obtained similar results for LULC when comparing conventional learning models (random forest) versus MLP at the fusion of Sentinel-1 and Sentinel-2. 
We have not yet found empirical evidence in EO that compares the increase in complexity in the NN models before the fusion versus the NN models after, which could be interesting to analyze and know which part to dedicate more resources to.
However, when focusing on the complexity of the view-encoder models there are more aspects of attention that needs to be set, such as regularization techniques including dropout, batch-normalization, pre-training, data-augmentation, and early-stopping.

\section{Additional components} \label{sec:addcomp}

In addition to previous questions and choices, some works have proposed different components or modules that help in learning (stability and predictive performance) MV models. In particular, the dropout\footnote{Regularization technique that is used to drop or sample some neurons on each layer of the neural network. The drop is performed based on a probability defined as a hyper-parameter.} \citep{srivastava2014dropout} has been mentioned as crucial technique to include throughout the MV model for better learning \citep{chen2017deepfusionremotea,zhang2020combiningopticalfluorescencea,maimaitijiang2020soybeanyieldpredictiona,rashkovetsky2021wildfiredetectionmultisensor}.
Other works \citep{chen2017deepfusionremotea,audebert2017semanticsegmentationeartha,xu2018multisourceremotesensing,hang2020classificationhyperspectrallidar,maimaitijiang2020soybeanyieldpredictiona,hong2021morediversemeansa,wu2021convolutionalneuralnetworks,rashkovetsky2021wildfiredetectionmultisensor,zhao2022multisourcecollaborativeenhanced} mentioned the batch-normalization\footnote{Technique used for reducing internal covariance-shift of the layers (and also used as regularization). It normalizes the features on each layer, allowing to reverse of the operation through two learnable parameters.} \citep{ioffe2015batch} with the same purpose. For example, \cite{wang2020deepmultimodalfusion} proposed to share parameters between view-predictive models but maintain the batch-normalization specific for each view in multi-sensor image segmentation.

\begin{figure}[t]
\includegraphics[width=0.95\textwidth, page=2]{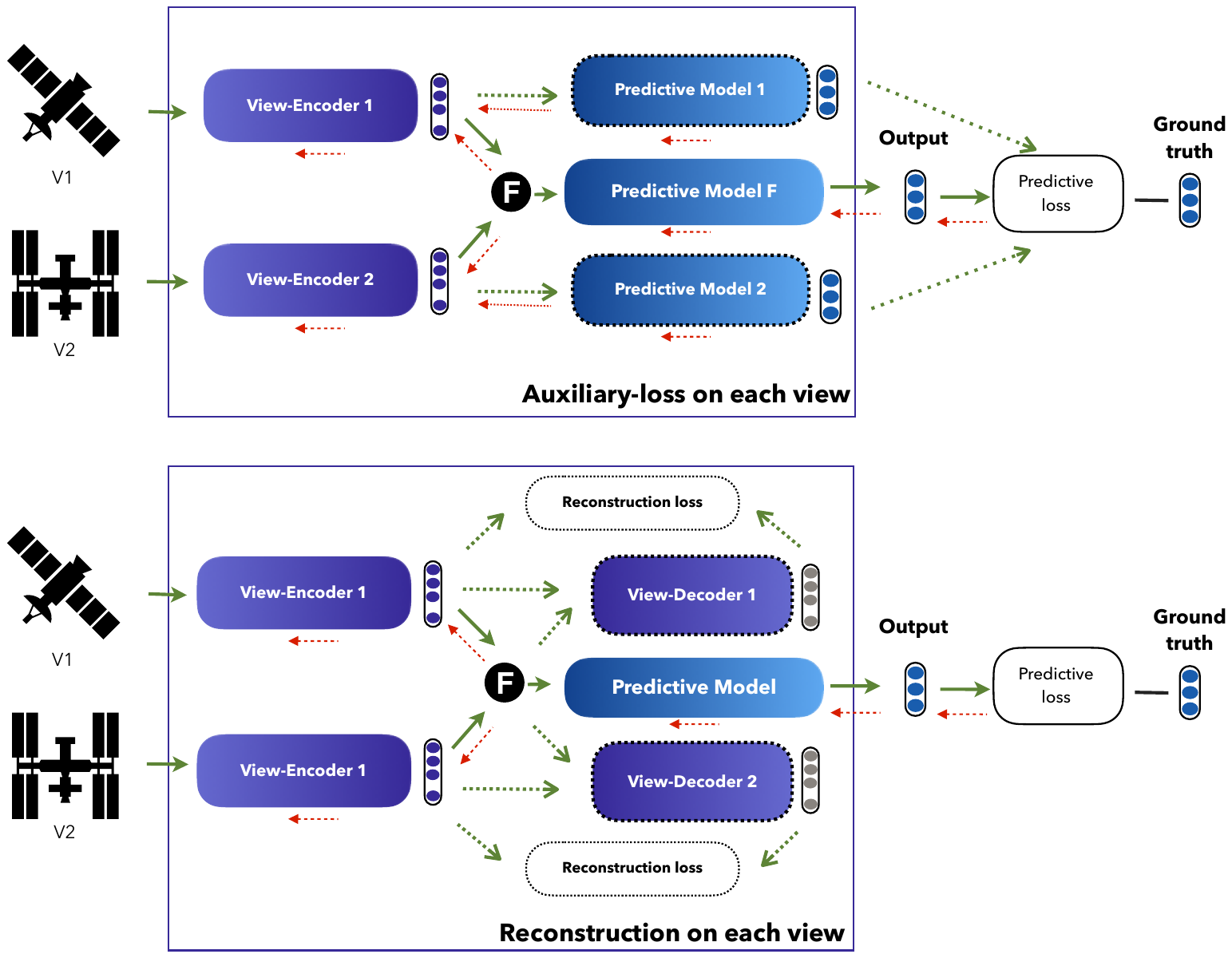}
\caption{Illustration of additional components in feature-level fusion: auxiliary-loss on each view at the top, and reconstruction on each view at the bottom. The forward pass of the model is from left to right (green arrows), while the backward pass is from right to left (red dashed arrows). The dashed green arrows are the auxiliary forward, only used for training. The ``preditice model F'' represent the predictive model that is fed with the fused representation.}\label{fig:fusion_types:extra2}
\end{figure}

Some techniques are used to act as a regularization, i.e. improve model generalization or reduce over-fitting (as well as inductive bias) are:
\begin{itemize}
    \item Feature reduction or selection \citep{debes2014hyperspectrallidardata,khodadadzadeh2015fusionhyperspectrallidar,bocca2016effecttuningfeaturea,ghamisi2017hyperspectrallidardata,denize2019evaluationusingsentinel1,kang2020comparativeassessmentenvironmentala}: this consist on reducing the number of features (or the number of bands in the case of images) in each view, focusing on removing the redundant or non-relevant information of the data. This approach prevents the model from suffering the curse of dimensionality \citep{kang2020comparativeassessmentenvironmentala} or ``collapse'' (diverge in learning). For instance, \cite{ghamisi2017hyperspectrallidardata} reduced the number of bands in optical-HS images with proposed extinction profiles and kernel principal component analysis. While other works use NNs to extract features, e.g. NN features for different bands of optical-MS images \citep{nijhawan2017deeplearninghybrid}, or for image and text \citep{mantsis2022multimodalfusionsentinel}.
    \item Share parameters \citep{nguyen2019spatialtemporalmultitasklearninga,srivastava2019understandingurbanlanduse,wang2020deepmultimodalfusion,liu2020multiviewselfconstructinggrapha,wang2021multiviewattentioncnnlstm,hang2020classificationhyperspectrallidar}: in order to avoid over-parameterization in the MV learning model, this technique uses the same layers with parameter sharing in the NN for each view, as SiameseNets (\citep{bromley1993signatureverificationusing}, see \citep{chicco2021siameseneuralnetworks} for a survey), or as called in more recent works, TwinNets \citep{zbontar2021barlowtwinsselfsupervised}. The sharing is usually applied in the whole NN for each view, but it has also been applied in just a few layers of 2D CNN that process optical-MS and SAR views \citep{cuelarosa2021investigatingfusionstrategiesa}.  
    Please note that this could only be used when the same NN architectures are designed for each view.
    
    \item Grouping \citep{audebert2017semanticsegmentationeartha,audebert2018rgbveryhigh,wang2020winterwheatyieldb,irvin2020forestnetclassifyingdrivers,cao2021integratingmultisourcedata, zheng2022gathertoguidenetworkremote}: also in order to reduce over-parametrization by having multiple models, this approach groups views based on their semantic or structural similarity. In practice, it considers the concatenation or stacking of the MV input data. For example, \cite{wang2020winterwheatyieldb} grouped temporal features (optical-MS and meteorological data) as one view, and static features (soil information) as another view for agricultural yield prediction. 
    Please note that this could be performed when the view-features are from the same data type (e.g. images, vectors).
    
    \item Use pre-trained NNs: as the name suggest, it re-uses networks that are pre-trained on large scale datasets. The main constraint is that the EO data has to have the same (or similar) resolution as the data that the pre-trained model was trained on, e.g. Imagenet-based CNNs need RGB (or three bands) images. However, some works have used multi-band images with the first layer initialized randomly or some heuristic followed by the layers of the pre-trained model \citep{rudner2019multi3netsegmentingflooded,irvin2020forestnetclassifyingdrivers,rambour2020flooddetectiontime,dimartino2021multibranchdeeplearning,yuan2021deeplearningbasedmultispectralsatellite,zhou2022cegfnetcommonextraction}.
    
    \item Prediction loss on each view \citep{benedetti2018textfusiondeeplearninga,ienco2019combiningsentinel1sentinel2b,cuelarosa2021investigatingfusionstrategiesa,hosseinpour2022cmgfnetdeepcrossmodala,hang2020classificationhyperspectrallidar,saintefaregarnot2022multimodaltemporalattention}: this consist on having an additional predictive loss for each view, forcing all views to be used in the downstream task. This is done by including one predictive model for each view instead of having a single model after the fusion, see Figure~\ref{fig:fusion_types:extra2} for an example. There are some differences in the proposal regarding how to apply weights to the additional MV losses:
    \begin{itemize}
        \item In some works \citep{benedetti2018textfusiondeeplearninga,hang2020classificationhyperspectrallidar} they decided to set different weights for the loss of each view in LULC, which aligns to the arguments that views converge at different rates \citep{wang2020whatmakestraininga}. \cite{wang2020whatmakestraininga} even proposed assigning dynamic weights during training, however, they noted that a fixed weight was enough in most cases.
        \item \cite{ienco2019combiningsentinel1sentinel2b} assigned one weight to the total sum of the additional MV losses,    
        \item while others decided not to use weights \citep{cuelarosa2021investigatingfusionstrategiesa,hosseinpour2022cmgfnetdeepcrossmodala}, just the sum of all the MV losses.
    \end{itemize}
    
    \item Reconstruct views \citep{wu2021convolutionalneuralnetworks, zhang2022informationfusionclassification}: which involves having additional losses focused on learning to reconstruct each view entirely \citep{zhang2022informationfusionclassification} or the view-representation previous to fusion \citep{wu2021convolutionalneuralnetworks}. The idea is that the joint representation should contain enough information for the reconstruction of individual views, see Figure~\ref{fig:fusion_types:extra2} for an example.
\end{itemize}

\unskip

Other techniques used with the purpose of giving stability to the training, i.e. help model convergence and learning, are mentioned in the following:
\begin{itemize}
    \item Perform pre-training and fine-tuning \citep{feichtenhofer2016convolutionaltwostreamnetworka,perez-rua2019mfasmultimodalfusion,wang2020whatmakestraininga}: this consist of doing the pre-training of each view-encoder individually on the same EO data that is used for the downstream task, in contrast to the previous technique commented that use already pre-trained models. For instance, learn to predict the downstream task based on each view individually \citep{xu2018multisourceremotesensing}, or learn to reconstruct the views as a pre-training \citep{zhang2022informationfusionclassification}, then fine-tune with the downstream task.
    With this, the parameters of the view-encoders are not randomly initialized but they already have information about the EO patterns of the used dataset.
    
    \item Multiple fusion channels \citep{hazirbas2017fusenetincorporatingdeptha,audebert2018rgbveryhigh,gadiraju2020multimodaldeeplearninga,hang2020classificationhyperspectrallidar,zhang2020hybridattentionawarefusion,cuelarosa2021investigatingfusionstrategiesa,cao2021integratingmultisourcedata,li2022outcome2021ieee,zhou2022cegfnetcommonextraction,zhao2022multisourcecollaborativeenhanced}: as already commented in previous sections, this applies fusion at different stages of the NN model (e.g. hybrid fusion), with the purpose that the information given to the fusion is at multiple levels of abstraction. However, it has been shown in the vision domain that fusion in just some layers is better than in all \citep{vielzeuf2018centralnetmultilayerapproach,perez-rua2019mfasmultimodalfusion}.
    
    \item Residual fusion learning \citep{audebert2017semanticsegmentationeartha,audebert2018rgbveryhigh,cao2021c3netcrossmodalfeaturea,zhang2022informationfusionclassification,hosseinpour2022cmgfnetdeepcrossmodala,zhang2022informationfusionclassification,zheng2022gathertoguidenetworkremote}: it consists of setting skip connections in the merge functions in order to learn residual after fusion, which is inspired by residual NN (or ResNet \citep{he2016deepresiduallearning}) and its ability to build very deep models.
    
    \item Normalize view-representations \citep{marmanis2016semanticsegmentationaerial,zhang2022informationfusionclassification}: this corresponds to the normalization of the view-representation (features) before the merge function is applied, ideally to handle the different feature scales  of the MV data. For example, \cite{marmanis2016semanticsegmentationaerial} applied it to optical (RG+NIR) and DSM view-representations, and \cite{wang2020deepmultimodalfusion} remarked in their proposal that having a specific normalization for each view was crucial.
    
    \item Different learning for views \citep{wang2020whatmakestraininga,zhang2020hybridattentionawarefusion}: motivated by the factor that views could have information at different levels, this technique assigns different learning rates for each view in the model optimization. To give an idea, \cite{zhang2020hybridattentionawarefusion} used a higher learning rate on DSM views and lower on optical-RGB views for LULC.
\end{itemize}

Other approaches have used post-processing after training, e.g. update the predictions based on the application and domain-knowledge \citep{nemmour2006multiplesupportvector,debes2014hyperspectrallidardata,robinson2021globallandcovermapping,ma2021outcome2021ieee,rashkovetsky2021wildfiredetectionmultisensor,li2022outcome2021ieee}.
Furthermore, some have included context information in the modeling, e.g. neighborhood pixels as input data on pixel-wise predictions \citep{chen2014deeplearningbasedclassification,debes2014hyperspectrallidardata,kussul2017deeplearningclassificationa,ghamisi2017hyperspectrallidardata,chen2017deepfusionremotea,nijhawan2017deeplearninghybrid,xu2018multisourceremotesensing,benedetti2018textfusiondeeplearninga,gadiraju2020multimodaldeeplearninga,wu2021convolutionalneuralnetworks,hong2021morediversemeansa,gao2022adaptivespectralspatialfeature} or graph-based constraints on similar pixels/patches  \citep{debes2014hyperspectrallidardata,ghamisi2017hyperspectrallidardata,liu2020multiviewselfconstructinggrapha,zhao2020jointclassificationhyperspectral,hong2021multimodalremotesensing}.

\section{Individual Views Analysis} \label{sec:indviews}
The following analysis provides a general perspective of the sources in the EO domain and focuses on the predictive performance of the downstream task. Please note that a similar analysis of views could be made based on explainability methods, however, we only focus on the predictive performance in this manuscript.

\subsection{Which Views are Most Used in Earth Observation?}

Optical (surface reflectance) data is the view that usually provides more information and, therefore, is the most used for various downstream EO tasks. Meanwhile, active-based views are the quintessential views chosen to complement (and improve by fusion) the optical in classification tasks with MV models, e.g. by using SAR view \citep{rudner2019multi3netsegmentingflooded,ienco2019combiningsentinel1sentinel2b,rambour2020flooddetectiontime,hong2021morediversemeansa,cuelarosa2021investigatingfusionstrategiesa,ofori-ampofo2021croptypemappingb,sebastianelli2021paradigmselectiondata,song2021evaluationlandsatsentinel2,dimartino2021multibranchdeeplearning,saintefaregarnot2022multimodaltemporalattention,mantsis2022multimodalfusionsentinel} or LiDAR view \citep{xu2018multisourceremotesensing,mohla2020fusatnetdualattention,hang2020classificationhyperspectrallidar,hong2021morediversemeansa,wu2021convolutionalneuralnetworks,zhang2022informationfusionclassification}.
Furthermore, the DSM view has been widely used together with the optical view
\citep{marmanis2016semanticsegmentationaerial,chen2017deepfusionremotea,nijhawan2017deeplearninghybrid,audebert2017semanticsegmentationeartha,audebert2018rgbveryhigh,liu2018deepconvolutionalneural,zhang2020hybridattentionawarefusion,maimaitijiang2020soybeanyieldpredictiona,irvin2020forestnetclassifyingdrivers,cao2021c3netcrossmodalfeaturea,hosseinpour2022cmgfnetdeepcrossmodala,zheng2022gathertoguidenetworkremote,zhou2022cegfnetcommonextraction,zhao2022multisourcecollaborativeenhanced}.
The visible light of the spectrum (RGB bands concretely) is more relevant than other spectral bands in the optical view with respect to the predictive performance of NN models \citep{sa2016deepfruitsfruitdetection,nevavuori2019cropyieldpredictiona,rambour2020flooddetectiontime}. Besides, views with coarse resolutions are usually worst in predictive performance than finer-resolution views \citep{rudner2019multi3netsegmentingflooded,gadiraju2020multimodaldeeplearninga,song2021evaluationlandsatsentinel2}, and that multi-temporal views (data with temporal information) got better predictive performance than static views \citep{gadiraju2020multimodaldeeplearninga,rambour2020flooddetectiontime,diaconu2022understandingroleweather}.
Although RS-based views provide valuable and more accessible information for downstream EO tasks than ground-based views \citep{srivastava2019understandingurbanlanduse,heidler2021selfsupervisedaudiovisualrepresentation}, some works have used intriguing data sources to complement RS-based views. For instance, \cite{heidler2021selfsupervisedaudiovisualrepresentation} used ground-based audios in addition to the optical view to classify an observed place, or \cite{mantsis2022multimodalfusionsentinel} which included images and text from tweets to estimate the snow depth of an observed place. These works suggest that social media could be a powerful source to estimate disasters \citep{said2019naturaldisastersdetection}, such as earthquakes \citep{cresci2017nowcastingearthquakeconsequences} and floods \citep{bischke2018multimediasatellitetask}.
There are other cases of domain-specific views depending on the application. 
One is the case of the agricultural yield prediction, where the weather and soil views are chosen to complement (and improve) the optical view \citep{nguyen2019spatialtemporalmultitasklearninga,cai2019integratingsatelliteclimatea,wang2020winterwheatyieldb,meng2021predictingmaizeyielda,cao2021integratingmultisourcedata}, or even used without the optical view \citep{khaki2020cnnrnnframeworkcropa}, in addition to different types of metadata, e.g. stats of the planted crop \citep{gangopadhyay2019deeptimeseries,shahhosseini2021cornyieldpredictiona}, the region where it was planted \citep{chu2020endtoendmodelricea}, or irrigation factors \citep{zhang2020combiningopticalfluorescencea, ahamed2015applyingdatamining}.

Some works include views from the same source, e.g. different time periods of the same sensor are taken as multi-input in the MV learning model to predict LULC changes across time \citep{nemmour2006multiplesupportvector}. 
While some works generate multiple views from single-view data, e.g. by different image operations (zoom ratios \citep{luus2015multiviewdeeplearning}, color alteration \citep{yang2017deepmultimodalrepresentation}, rotations \citep{liu2020multiviewselfconstructinggrapha}). 
Others, partition the single-view data to generate MV, e.g. split the spectral bands of Sentinel-2 optical image for vegetation recognition \citep{nijhawan2017deeplearninghybrid} or water body detection \citep{yuan2021deeplearningbasedmultispectralsatellite}. Furthermore, feature extraction of spectral indexes  \citep{audebert2017semanticsegmentationeartha,audebert2018rgbveryhigh,denize2019evaluationusingsentinel1,ienco2019combiningsentinel1sentinel2b,nguyen2019spatialtemporalmultitasklearninga}, spectral degradation \citep{hong2021multimodalremotesensing}, or model-based (such as PCA \citep{gao2022adaptivespectralspatialfeature}) have been used as an additional view to the raw bands of optical images. This shows that different types of views can be considered for different predictive tasks and that ``view" is a concept that can allow greater flexibility and exploration of methods in the same framework.

\subsection{Does the use of additional views improve prediction performance?} 
Outside the EO domain, substantial evidence suggests that additional views or modalities improve the predictive performance on downstream tasks with respect to single-view data \citep{georgiou2019deephierarchicalfusion,wang2019deepmultiviewinformationa,perez-rua2019mfasmultimodalfusion,arevalo2020gatedmultimodalnetworksa}. However, does this result apply to EO applications? Below we gathered some results in this direction.

There is plenty of evidence from works that use two-view data on various EO tasks, showing that the predictive performance improves with respect to training on any of the single-views, e.g. with optical and active-based views (SAR/LiDAR/DSM) \citep{debes2014hyperspectrallidardata,chen2017deepfusionremotea,xu2018multisourceremotesensing,audebert2018rgbveryhigh,ienco2019combiningsentinel1sentinel2b,rambour2020flooddetectiontime,mohla2020fusatnetdualattention,hang2020classificationhyperspectrallidar,zhang2020hybridattentionawarefusion,ofori-ampofo2021croptypemappingb,cuelarosa2021investigatingfusionstrategiesa,hong2021morediversemeansa,hong2021multimodalremotesensing,konapala2021exploringsentinel1sentinel2,wu2021convolutionalneuralnetworks,konapala2021exploringsentinel1sentinel2,hosseinpour2022cmgfnetdeepcrossmodala,zheng2022gathertoguidenetworkremote}. This indicates that the views complement each other in the MV learning for EO tasks, and happen when other types of view are chosen to supplement or replace the optical view \citep{sa2016deepfruitsfruitdetection,srivastava2019understandingurbanlanduse,gadiraju2020multimodaldeeplearninga,chu2020endtoendmodelricea,rashkovetsky2021wildfiredetectionmultisensor,shahhosseini2021cornyieldpredictiona}.
Several publications show that the improvements with respect to the optical view, appear even when using more than two views. 
\cite{nguyen2019spatialtemporalmultitasklearninga} demonstrated that optical-MS images, NVDI, and soil properties perform better than individual views for agricultural yield prediction. \cite{pageot2020detectionirrigatedrainfed} showed that optical-MS image, SAR image, and weather improve over individual views for irrigation recognition. \cite{song2021evaluationlandsatsentinel2} fused optical-MS from Sentinel-2, optical from MODIS, optical from Landsat, and SAR from Sentinel-1 views and find better predictive performance as compared to individual views for crop-type mapping. \cite{irvin2020forestnetclassifyingdrivers} showed that optical and auxiliary data extracted from DEM, weather, and soil properties improve over optical for detecting deforestation. In \cite{luus2015multiviewdeeplearning} work, multiple zooms of an optical image improved predictive performance over a single perspective for LULC. \cite{wang2021multiviewattentioncnnlstm} found that multiple angles of an object improve over a single angle for automatic target recognition.
Finally, some works \citep{nijhawan2017deeplearninghybrid,pei2018sarautomatictarget,cai2019integratingsatelliteclimatea,rudner2019multi3netsegmentingflooded,wang2020winterwheatyieldb,zhang2020combiningopticalfluorescencea,pageot2020detectionirrigatedrainfed,maimaitijiang2020soybeanyieldpredictiona,diaconu2022understandingroleweather} presented a monotonically increase in predictive performance by including additional views in the MV model. 
This evidence suggests that learning from MV data outperforms in terms of predictive performance as compared to learning from single-view data in the EO domain. 
\cite{heidler2021selfsupervisedaudiovisualrepresentation} showed an interesting result that MV learning improves over single-view learning when predicting with single-view data. They showed that a MV model trained with optical and audio views outperforms a single-view trained model when predicting using only the optical view. This suggests that additional views could benefit the model even if they are not used during test time. 

Even though all the previously commented works report that using additional data improves predictive performance, there are some works that report the opposite referring to the number of features in single-view learning with conventional machine learning models. 
In these works \citep{chen2014deeplearningbasedclassification,bocca2016effecttuningfeaturea,kang2020comparativeassessmentenvironmentala}, a subset of the features (selected with feature engineering techniques) improved the predictive performance of the models. This might suggest that conventional models with single-view learning fail to adequately extract the patterns contained in the additional views needed for the downstream task.
However, it is important to note that the results usually depend on the empirical data, as well as on the downstream task \citep{song2021evaluationlandsatsentinel2}, the difficulty of that task (e.g. number of classes, type of task or granularity) \citep{khaki2020cnnrnnframeworkcropa}, the number of training examples \citep{wang2018deeptransferlearninga,tseng2021cropharvestglobaldataset,stojnic2021selfsupervisedlearningremote}, or the model used \citep{denize2019evaluationusingsentinel1}.

\section{Conclusion and Outlook} \label{sec:concl}

Although many approaches have been explored in the literature on data fusion in MV learning, there are still some open challenges that could motivate new research and proposals in the EO domain.
\begin{enumerate}
    \item The usual assumption of MV learning is that all views are available for each sample during and after training. However, EO scenarios are dynamic environments that do not necessarily follow this assumption, e.g. remote sensors, may fail or be unavailable, causing a MV learning with missing views \citep{shen2015missinginformationreconstruction}. Only a few have explored the effect of fusion when this occurs. For instance, fusions further away from the input data (e.g. decision and feature fusion) are more robust to missing\footnote{In practice the missing is filled with zero when feeding the model.} in HS and LiDAR images \citep{hong2021morediversemeansa}. Other works evidence the same result when missing\footnote{In practice the missing is discarded or ignored by the model.} optical-MS images at some time-steps of an optical and SAR MV model \citep{saintefaregarnot2022multimodaltemporalattention,ofori-ampofo2021croptypemappingb} or in cloudy weathers \citep{yuan2021deeplearningbasedmultispectralsatellite,rashkovetsky2021wildfiredetectionmultisensor}.  
    \cite{hong2021morediversemeansa} also showed that the robustness to missing views could be increased by including additional components to the standard fusion approaches (as the ones in Figure~\ref{fig:fusion_types}).
    These solutions are usually data specific and require knowing in advance what and when the missing views will be. However, NN models have shown a great ability to reconstruct complex EO patterns \citep{diaconu2022understandingroleweather}.
    The problem of how to adapt the MV learning model to missing views still has open points that could motivate further research. For example, consider when more than two views are used, the missing views can be dynamic and a robust model could be suitable.
    
    \item There are not many studies about the uncertainty of the prediction when using MV data. It is reasonable to ask whether additional views reduce or increase the uncertainty. It might happen that if views are too different from each other, the uncertainty increases, or if views are more similar to each other, the uncertainty is reduced. \cite{ofori-ampofo2021croptypemappingb} showed that when executing the model multiple times the variance gets reduced by using MV models instead of the single-view ones.
    However, it is not clear what is the expected effect of uncertainty in MV learning for EO.
    
    \item Multiple works in the EO domain have focused on the proposal of different models and architectures for the MV learning \citep{dealwis2019duoattentiondeep,marmanis2016semanticsegmentationaerial,ienco2019combiningsentinel1sentinel2b,cao2021c3netcrossmodalfeaturea,hosseinpour2022cmgfnetdeepcrossmodala,zhang2022informationfusionclassification,zhao2022multisourcecollaborativeenhanced,zhou2022cegfnetcommonextraction}, e.g by making them more complex and thereby enable them to extract more information. However, the usual approach is to select a fixed merge function (e.g. concatenation) and experiment with different types of NN architectures. Therefore, there is not much empirical comparison of different ways to fuse the view-representations, leaving it as an open problem.
    
    \item The explainability of a single-view model is a research line by itself with different perspectives. What is the impact on the explainability or interpretability as the complexity of the model increases by including MV data? Since current  approaches to deal with MV data is to use multiple models for each view, complexity is increased for each view.  It is challenging to understand what the model is doing under abstract operations and modules, therefore significant attention needs to be put at this point.
\end{enumerate}

This manuscript analyses different aspects of data fusion in MV learning in the context of EO data sources. Different approaches from the literature are grouped based on their similarity, providing a unified structure to compare and share results in the EO domain.
To the best of our knowledge, the works included in this review reflect the current trends of MV fusion learning in the EO domain.

\section*{Abbreviations}
The following abbreviations are used in this manuscript:\\
\noindent 
\begin{tabular}{ll}
CNN & Convolutional Neural Network \\
DEM & Digital Elevation Model \\
DTM & Digital Terrain Model \\
DSM & Digital Surface Model \\
EO & Earth Observation \\
EVI & Enhanced Vegetation Index \\
GRU & Gated Recurrent Unit \\
HS & Hyper-Spectral \\
LiDAR & Light Detection And Ranging \\
LSTM & Long-Short Term Memory \\
LULC & Land-Use Land-Cover \\
MODIS & Moderate Resolution Imaging Spectroradiometer \\
MS & Multi-Spectral \\
MLP & Multi-Layer Perceptron \\
MV & Multi-View \\
NIR & Near Infra-Red \\
NDVI & Normalized Difference Vegetation Index \\
NN & Neural Network \\
PCA & Principal Component Analysis \\
RS & Remote Sensing \\
RGB & Red-Green-Blue \\
RNN & Recurrent Neural Network \\
SAR & Synthetic Aperture Radar \\
UAV & Unmanned Aerial Vehicles \\
\end{tabular}

\section*{CRediT authorship contribution statement}
Francisco Mena: Conceptualization, methodology, investigation, writing---original draft, visualization; 
Diego Arenas: Conceptualization, methodology, writing---review and editing, supervision; 
Marlon Nuske: Conceptualization, writing---review and editing, project administration;
Andreas Dengel: Supervision, funding acquisition, resources.

\section*{Data availability}
No new dataset were created or analyzed in this study.

\section*{Acknowledgments}
F. Mena acknowledges the financial support from the chair of Prof. Dr. Prof. h.c. Andreas Dengel with Technische Universität Kaiserslautern (TUK).

\bibliographystyle{plainnat}
\bibliography{content/refs}

\begin{thebibliography}{170}
\providecommand{\natexlab}[1]{#1}
\providecommand{\url}[1]{\texttt{#1}}
\expandafter\ifx\csname urlstyle\endcsname\relax
  \providecommand{\doi}[1]{doi: #1}\else
  \providecommand{\doi}{doi: \begingroup \urlstyle{rm}\Url}\fi

\bibitem[Ahamed et~al.(2015)Ahamed, Mahmood, Hossain, Kabir, Das, Rahman, and
  Rahman]{ahamed2015applyingdatamining}
A.~T. M~Shakil Ahamed, Navid~Tanzeem Mahmood, Nazmul Hossain, Mohammad~Tanzir
  Kabir, Kallal Das, Faridur Rahman, and Rashedur~M Rahman.
\newblock Applying data mining techniques to predict annual yield of major
  crops and recommend planting different crops in different districts in
  {{Bangladesh}}.
\newblock In \emph{{{International Conference}} on {{Software Engineering}},
  {{Artificial Intelligence}}, {{Networking}} and {{Parallel}}/{{Distributed
  Computing}} ({{SNPD}})}, pages 1--6, 2015.
\newblock \doi{10.1109/SNPD.2015.7176185}.

\bibitem[Ahmad et~al.(2017)Ahmad, Pogorelov, Riegler, Conci, and
  Halvorsen]{ahmad2017cnnganbased}
Kashif Ahmad, Konstantin Pogorelov, Michael Riegler, Nicola Conci, and Pål
  Halvorsen.
\newblock \emph{{{CNN}} and {{GAN Based Satellite}} and {{Social Media Data
  Fusion}} for {{Disaster Detection}}}.
\newblock 2017.

\bibitem[Albughdadi et~al.(2017)Albughdadi, Kouam{\'e}, Rieu, and
  Tourneret]{albughdadi2017missingdatareconstruction}
Mohanad Albughdadi, Denis Kouam{\'e}, Guillaume Rieu, and Jean-Yves Tourneret.
\newblock Missing data reconstruction and anomaly detection in crop development
  using agronomic indicators derived from multispectral satellite images.
\newblock In \emph{IEEE International Geoscience and Remote Sensing Symposium
  (IGARSS)}, pages 5081--5084. IEEE, 2017.
\newblock \doi{10.1109/IGARSS.2017.8128145}.

\bibitem[Andrew et~al.(2013)Andrew, Arora, Bilmes, and
  Livescu]{andrew2013deepcanonicalcorrelationa}
Galen Andrew, Raman Arora, Jeff Bilmes, and Karen Livescu.
\newblock Deep {{Canonical Correlation Analysis}}.
\newblock In \emph{Proceedings of {{International Conference}} on {{Machine
  Learning}} (ICML)}, pages 1247--1255. {PMLR}, 2013.

\bibitem[Arevalo et~al.(2017)Arevalo, Solorio, Montes-y Gómez, and
  González]{arevalo2017gatedmultimodalunits}
John Arevalo, Thamar Solorio, Manuel Montes-y Gómez, and Fabio~A. González.
\newblock Gated {{Multimodal Units}} for {{Information Fusion}}.
\newblock \emph{International Conference on Learning Representation (ICLR)
  Workshop}, 2017.

\bibitem[Arevalo et~al.(2020)Arevalo, Solorio, Montes-y Gómez, and
  González]{arevalo2020gatedmultimodalnetworksa}
John Arevalo, Thamar Solorio, Manuel Montes-y Gómez, and Fabio~A. González.
\newblock Gated multimodal networks.
\newblock \emph{Neural Computing and Applications}, 32\penalty0 (14):\penalty0
  10209--10228, 2020.
\newblock ISSN 1433-3058.
\newblock \doi{10.1007/s00521-019-04559-1}.

\bibitem[Audebert et~al.(2017)Audebert, Le~Saux, and
  Lefèvre]{audebert2017semanticsegmentationeartha}
Nicolas Audebert, Bertrand Le~Saux, and Sébastien Lefèvre.
\newblock Semantic {{Segmentation}} of {{Earth Observation Data Using
  Multimodal}} and {{Multi-scale Deep Networks}}.
\newblock In \emph{Asian Conference on Computer Vision (ACCV)}, pages 180--196.
  {Springer International Publishing}, 2017.
\newblock ISBN 978-3-319-54181-5.
\newblock \doi{10.1007/978-3-319-54181-5_12}.

\bibitem[Audebert et~al.(2018)Audebert, Le~Saux, and
  Lefèvre]{audebert2018rgbveryhigh}
Nicolas Audebert, Bertrand Le~Saux, and Sébastien Lefèvre.
\newblock Beyond {{RGB}}: {{Very}} high resolution urban remote sensing with
  multimodal deep networks.
\newblock \emph{ISPRS Journal of Photogrammetry and Remote Sensing},
  140:\penalty0 20--32, 2018.
\newblock ISSN 0924-2716.
\newblock \doi{10.1016/j.isprsjprs.2017.11.011}.

\bibitem[Bargoti and Underwood(2017)]{bargoti2017deepfruitdetection}
Suchet Bargoti and James Underwood.
\newblock Deep fruit detection in orchards.
\newblock In \emph{{{IEEE International Conference}} on {{Robotics}} and
  {{Automation}} ({{ICRA}})}, pages 3626--3633, 2017.
\newblock \doi{10.1109/ICRA.2017.7989417}.

\bibitem[Benedetti et~al.(2018)Benedetti, Ienco, Gaetano, Ose, Pensa, and
  Dupuy]{benedetti2018textfusiondeeplearninga}
Paola Benedetti, Dino Ienco, Raffaele Gaetano, Kenji Ose, Ruggero~G. Pensa, and
  Stephane Dupuy.
\newblock \${{M}}\^3\textbackslash{{textFusion}}\$: {{A Deep Learning
  Architecture}} for {{Multiscale Multimodal Multitemporal Satellite Data
  Fusion}}.
\newblock \emph{IEEE Journal of Selected Topics in Applied Earth Observations
  and Remote Sensing}, 11\penalty0 (12):\penalty0 4939--4949, 2018.
\newblock ISSN 2151-1535.
\newblock \doi{10.1109/JSTARS.2018.2876357}.

\bibitem[Benediktsson et~al.(1989)Benediktsson, Swain, and
  Ersoy]{benediktsson1989neuralnetworkapproaches}
J.A. Benediktsson, P.H. Swain, and O.K. Ersoy.
\newblock Neural {{Network Approaches Versus Statistical Methods}} in
  {{Classification}} of {{Multisource Remote Sensing Data}}.
\newblock In \emph{{{Canadian Symposium}} on {{Remote Sensing Geoscience}} and
  {{Remote Sensing Symposium}}}, volume~2, pages 489--492, 1989.
\newblock \doi{10.1109/IGARSS.1989.578748}.

\bibitem[Bhojani and Bhatt(2020)]{bhojani2020wheatcropyield}
Shital~H. Bhojani and Nirav Bhatt.
\newblock Wheat crop yield prediction using new activation functions in neural
  network.
\newblock \emph{Neural Computing and Applications}, 32\penalty0 (17):\penalty0
  13941--13951, 2020.
\newblock ISSN 1433-3058.
\newblock \doi{10.1007/s00521-020-04797-8}.

\bibitem[Bischke et~al.(2018)Bischke, Helber, Zhao, de~Bruijn, and
  Borth]{bischke2018multimediasatellitetask}
Benjamin Bischke, Patrick Helber, Zhengyu Zhao, Jens de~Bruijn, and Damian
  Borth.
\newblock \emph{The {{Multimedia Satellite Task}} at {{MediaEval}} 2018
  {{Emergency Response}} for {{Flooding Events}}}.
\newblock 2018.

\bibitem[Bleiholder and Naumann(2009)]{bleiholder2009datafusion}
Jens Bleiholder and Felix Naumann.
\newblock Data fusion.
\newblock \emph{ACM Computing Surveys}, 41\penalty0 (1):\penalty0 1:1--1:41,
  2009.
\newblock ISSN 0360-0300.
\newblock \doi{10.1145/1456650.1456651}.

\bibitem[Blum and Mitchell(1998)]{blum1998combininglabeledunlabeled}
Avrim Blum and Tom Mitchell.
\newblock Combining labeled and unlabeled data with co-training.
\newblock In \emph{Proceedings of the Eleventh Annual Conference on
  {{Computational}} Learning Theory}, {{COLT}}' 98, pages 92--100. {Association
  for Computing Machinery}, 1998.
\newblock ISBN 978-1-58113-057-7.
\newblock \doi{10.1145/279943.279962}.

\bibitem[Bocca and Rodrigues(2016)]{bocca2016effecttuningfeaturea}
Felipe~F. Bocca and Luiz Henrique~Antunes Rodrigues.
\newblock The effect of tuning, feature engineering, and feature selection in
  data mining applied to rainfed sugarcane yield modelling.
\newblock \emph{Computers and Electronics in Agriculture}, 128:\penalty0
  67--76, 2016.
\newblock ISSN 0168-1699.
\newblock \doi{10.1016/j.compag.2016.08.015}.

\bibitem[Bosilj et~al.(2020)Bosilj, Aptoula, Duckett, and
  Cielniak]{bosilj2020transferlearningcropa}
Petra Bosilj, Erchan Aptoula, Tom Duckett, and Grzegorz Cielniak.
\newblock Transfer learning between crop types for semantic segmentation of
  crops versus weeds in precision agriculture.
\newblock \emph{Journal of Field Robotics}, 37\penalty0 (1):\penalty0 7--19,
  2020.
\newblock ISSN 1556-4967.
\newblock \doi{10.1002/rob.21869}.

\bibitem[Bromley et~al.(1993)Bromley, Guyon, LeCun, Säckinger, and
  Shah]{bromley1993signatureverificationusing}
Jane Bromley, Isabelle Guyon, Yann LeCun, Eduard Säckinger, and Roopak Shah.
\newblock Signature {{Verification}} using a "{{Siamese}}" {{Time Delay Neural
  Network}}.
\newblock \emph{Advances in {{Neural Information Processing Systems}} (NIPS)},
  6, 1993.

\bibitem[Cai et~al.(2019)Cai, Guan, Lobell, Potgieter, Wang, Peng, Xu, Asseng,
  Zhang, You, and Peng]{cai2019integratingsatelliteclimatea}
Yaping Cai, Kaiyu Guan, David Lobell, Andries~B. Potgieter, Shaowen Wang, Jian
  Peng, Tianfang Xu, Senthold Asseng, Yongguang Zhang, Liangzhi You, and Bin
  Peng.
\newblock Integrating satellite and climate data to predict wheat yield in
  {{Australia}} using machine learning approaches.
\newblock \emph{Agricultural and Forest Meteorology}, 274:\penalty0 144--159,
  2019.
\newblock ISSN 0168-1923.
\newblock \doi{10.1016/j.agrformet.2019.03.010}.

\bibitem[Camps-Valls et~al.(2008)Camps-Valls, Gomez-Chova, Munoz-Mari,
  Rojo-Alvarez, and
  Martinez-Ramon]{camps-valls2008kernelbasedframeworkmultitemporal}
Gustavo Camps-Valls, Luis Gomez-Chova, Jordi Munoz-Mari, JosÉ~Luis
  Rojo-Alvarez, and Manel Martinez-Ramon.
\newblock Kernel-{{Based Framework}} for {{Multitemporal}} and {{Multisource
  Remote Sensing Data Classification}} and {{Change Detection}}.
\newblock \emph{IEEE Transactions on Geoscience and Remote Sensing},
  46\penalty0 (6):\penalty0 1822--1835, 2008.
\newblock ISSN 1558-0644.
\newblock \doi{10.1109/TGRS.2008.916201}.

\bibitem[Camps-Valls et~al.(2021)Camps-Valls, Tuia, Zhu, and
  Reichstein]{camps-valls2021deeplearningearth}
Gustavo Camps-Valls, Devis Tuia, Xiao~Xiang Zhu, and Markus Reichstein.
\newblock \emph{Deep {{Learning}} for the {{Earth Sciences}}: {{A Comprehensive
  Approach}} to {{Remote Sensing}}, {{Climate Science}}, and {{Geosciences}}}.
\newblock {Wiley}, 2021.
\newblock ISBN 978-1-119-64614-3 978-1-119-64618-1.
\newblock \doi{10.1002/9781119646181}.

\bibitem[Cao et~al.(2021{\natexlab{a}})Cao, Zhang, Tao, Zhang, Luo, Zhang, Han,
  and Xie]{cao2021integratingmultisourcedata}
Juan Cao, Zhao Zhang, Fulu Tao, Liangliang Zhang, Yuchuan Luo, Jing Zhang,
  Jichong Han, and Jun Xie.
\newblock Integrating {{Multi-Source Data}} for {{Rice Yield Prediction}}
  across {{China}} using {{Machine Learning}} and {{Deep Learning Approaches}}.
\newblock \emph{Agricultural and Forest Meteorology}, 297:\penalty0 108275,
  2021{\natexlab{a}}.
\newblock ISSN 0168-1923.
\newblock \doi{10.1016/j.agrformet.2020.108275}.

\bibitem[Cao et~al.(2021{\natexlab{b}})Cao, Diao, Sun, Lyu, Yan, and
  Fu]{cao2021c3netcrossmodalfeaturea}
Zhiying Cao, Wenhui Diao, Xian Sun, Xiaode Lyu, Menglong Yan, and Kun Fu.
\newblock {{C3Net}}: {{Cross-Modal Feature Recalibrated}}, {{Cross-Scale
  Semantic Aggregated}} and {{Compact Network}} for {{Semantic Segmentation}}
  of {{Multi-Modal High-Resolution Aerial Images}}.
\newblock \emph{Remote Sensing}, 13\penalty0 (3):\penalty0 528,
  2021{\natexlab{b}}.
\newblock ISSN 2072-4292.
\newblock \doi{10.3390/rs13030528}.

\bibitem[Castanedo(2013)]{castanedo2013reviewdatafusion}
Federico Castanedo.
\newblock A {{Review}} of {{Data Fusion Techniques}}.
\newblock \emph{The Scientific World Journal}, 2013:\penalty0 e704504, 2013.
\newblock ISSN 2356-6140.
\newblock \doi{10.1155/2013/704504}.

\bibitem[Chen et~al.(2020{\natexlab{a}})Chen, Kornblith, Norouzi, and
  Hinton]{chen2020simpleframeworkcontrastive}
Ting Chen, Simon Kornblith, Mohammad Norouzi, and Geoffrey Hinton.
\newblock A {{Simple Framework}} for {{Contrastive Learning}} of {{Visual
  Representations}}.
\newblock In \emph{Proceedings of {{International Conference}} on {{Machine
  Learning}} (ICML)}, pages 1597--1607. {PMLR}, 2020{\natexlab{a}}.

\bibitem[Chen et~al.(2020{\natexlab{b}})Chen, Lu, and
  Wang]{chen2020deepcrossmodalimage}
Yaxiong Chen, Xiaoqiang Lu, and Shuai Wang.
\newblock Deep {{Cross-Modal Image}}–{{Voice Retrieval}} in {{Remote
  Sensing}}.
\newblock \emph{IEEE Transactions on Geoscience and Remote Sensing},
  58\penalty0 (10):\penalty0 7049--7061, 2020{\natexlab{b}}.
\newblock ISSN 1558-0644.
\newblock \doi{10.1109/TGRS.2020.2979273}.

\bibitem[Chen et~al.(2014)Chen, Lin, Zhao, Wang, and
  Gu]{chen2014deeplearningbasedclassification}
Yushi Chen, Zhouhan Lin, Xing Zhao, Gang Wang, and Yanfeng Gu.
\newblock Deep {{Learning-Based Classification}} of {{Hyperspectral Data}}.
\newblock \emph{IEEE Journal of Selected Topics in Applied Earth Observations
  and Remote Sensing}, 7\penalty0 (6):\penalty0 2094--2107, 2014.
\newblock ISSN 2151-1535.
\newblock \doi{10.1109/JSTARS.2014.2329330}.

\bibitem[Chen et~al.(2017)Chen, Li, Ghamisi, Jia, and
  Gu]{chen2017deepfusionremotea}
Yushi Chen, Chunyang Li, Pedram Ghamisi, Xiuping Jia, and Yanfeng Gu.
\newblock Deep {{Fusion}} of {{Remote Sensing Data}} for {{Accurate
  Classification}}.
\newblock \emph{IEEE Geoscience and Remote Sensing Letters}, 14\penalty0
  (8):\penalty0 1253--1257, 2017.
\newblock ISSN 1558-0571.
\newblock \doi{10.1109/LGRS.2017.2704625}.

\bibitem[Cheng et~al.(2021)Cheng, Zhou, Fu, Xu, and
  Zhang]{cheng2021deepsemanticalignment}
Qimin Cheng, Yuzhuo Zhou, Peng Fu, Yuan Xu, and Liang Zhang.
\newblock A {{Deep Semantic Alignment Network}} for the {{Cross-Modal
  Image-Text Retrieval}} in {{Remote Sensing}}.
\newblock \emph{IEEE Journal of Selected Topics in Applied Earth Observations
  and Remote Sensing}, 14:\penalty0 4284--4297, 2021.
\newblock ISSN 2151-1535.
\newblock \doi{10.1109/JSTARS.2021.3070872}.

\bibitem[Chew et~al.(2020)Chew, Rineer, Beach, O’Neil, Ujeneza, Lapidus,
  Miano, Hegarty-Craver, Polly, and Temple]{chew2020deepneuralnetworksa}
Robert Chew, Jay Rineer, Robert Beach, Maggie O’Neil, Noel Ujeneza, Daniel
  Lapidus, Thomas Miano, Meghan Hegarty-Craver, Jason Polly, and Dorota~S.
  Temple.
\newblock Deep {{Neural Networks}} and {{Transfer Learning}} for {{Food Crop
  Identification}} in {{UAV Images}}.
\newblock \emph{Drones}, 4\penalty0 (1):\penalty0 7, 2020.
\newblock ISSN 2504-446X.
\newblock \doi{10.3390/drones4010007}.

\bibitem[Chicco(2021)]{chicco2021siameseneuralnetworks}
Davide Chicco.
\newblock Siamese {{Neural Networks}}: {{An Overview}}.
\newblock In \emph{Artificial {{Neural Networks}}}, Methods in {{Molecular
  Biology}}, pages 73--94. {Springer US}, 2021.
\newblock ISBN 978-1-07-160826-5.
\newblock \doi{10.1007/978-1-0716-0826-5_3}.

\bibitem[Christoudias et~al.(2008)Christoudias, Urtasun, and
  Darrell]{christoudias2012multiviewlearningpresence}
C~Mario Christoudias, Raquel Urtasun, and Trevor Darrell.
\newblock Multi-{{View Learning}} in the {{Presence}} of {{View Disagreement}}.
\newblock In \emph{Proceedings of Conference on Uncertainty in Artificial
  Intelligence (UAI)}, pages 88--96, 2008.

\bibitem[Chu and Yu(2020)]{chu2020endtoendmodelricea}
Zheng Chu and Jiong Yu.
\newblock An end-to-end model for rice yield prediction using deep learning
  fusion.
\newblock \emph{Computers and Electronics in Agriculture}, 174:\penalty0
  105471, 2020.
\newblock ISSN 0168-1699.
\newblock \doi{10.1016/j.compag.2020.105471}.

\bibitem[Cresci et~al.(2017)Cresci, Avvenuti, La~Polla, Meletti, and
  Tesconi]{cresci2017nowcastingearthquakeconsequences}
Stefano Cresci, Marco Avvenuti, Mariantonietta La~Polla, Carlo Meletti, and
  Maurizio Tesconi.
\newblock Nowcasting of {{Earthquake Consequences}} using {{Big Social Data}}.
\newblock \emph{IEEE Internet Computing}, pages 1--1, 2017.
\newblock ISSN 1941-0131.
\newblock \doi{10.1109/MIC.2017.265102211}.

\bibitem[Cué La~Rosa et~al.(2021)Cué La~Rosa, Oliveira, and
  Feitosa]{cuelarosa2021investigatingfusionstrategiesa}
L.~E. Cué La~Rosa, D.~A.~B. Oliveira, and R.~Q. Feitosa.
\newblock Investigating {{Fusion Strategies}} on {{Encoder-Decoder Networks}}
  for {{Crop Segmentation Using SAR}} and {{Optical Image Sequences}}.
\newblock In \emph{{{IEEE International Geoscience}} and {{Remote Sensing
  Symposium}} (IGARSS)}, pages 2405--2408, 2021.
\newblock \doi{10.1109/IGARSS47720.2021.9554011}.

\bibitem[De~Alwis et~al.(2019)De~Alwis, Zhang, Na, and
  Li]{dealwis2019duoattentiondeep}
Sandya De~Alwis, Yishuo Zhang, Myung Na, and Gang Li.
\newblock Duo {{Attention}} with {{Deep Learning}} on {{Tomato Yield
  Prediction}} and {{Factor Interpretation}}.
\newblock In \emph{Pacific Rim International Conference on Artificial
  Intelligence}, Lecture {{Notes}} in {{Computer Science}}, pages 704--715.
  {Springer International Publishing}, 2019.
\newblock ISBN 978-3-030-29894-4.
\newblock \doi{10.1007/978-3-030-29894-4_56}.

\bibitem[Debes et~al.(2017)Debes, Merentitis, Heremans, Hahn, Frangiadakis, van
  Kasteren, Liao, Bellens, Pižurica, Gautama, Philips, Prasad, Du, and
  Pacifici]{debes2014hyperspectrallidardata}
Christian Debes, Andreas Merentitis, Roel Heremans, Jürgen Hahn, Nikolaos
  Frangiadakis, Tim van Kasteren, Wenzhi Liao, Rik Bellens, Aleksandra
  Pižurica, Sidharta Gautama, Wilfried Philips, Saurabh Prasad, Qian Du, and
  Fabio Pacifici.
\newblock Hyperspectral and {{LiDAR Data Fusion}}: {{Outcome}} of the 2013
  {{GRSS Data Fusion Contest}}.
\newblock \emph{IEEE Journal of Selected Topics in Applied Earth Observations
  and Remote Sensing}, 7\penalty0 (6):\penalty0 2405--2418, 2017.
\newblock ISSN 2151-1535.
\newblock \doi{10.1109/JSTARS.2014.2305441}.

\bibitem[Denize et~al.(2019)Denize, Hubert-Moy, Betbeder, Corgne, Baudry, and
  Pottier]{denize2019evaluationusingsentinel1}
Julien Denize, Laurence Hubert-Moy, Julie Betbeder, Samuel Corgne, Jacques
  Baudry, and Eric Pottier.
\newblock Evaluation of {{Using Sentinel-1}} and -2 {{Time-Series}} to
  {{Identify Winter Land Use}} in {{Agricultural Landscapes}}.
\newblock \emph{Remote Sensing}, 11\penalty0 (1):\penalty0 37, 2019.
\newblock ISSN 2072-4292.
\newblock \doi{10.3390/rs11010037}.

\bibitem[Di~Martino et~al.(2021)Di~Martino, Lenormand, and
  Koeniguer]{dimartino2021multibranchdeeplearning}
Thomas Di~Martino, Maxime Lenormand, and Elise~Colin Koeniguer.
\newblock Multi-{{Branch Deep Learning Model}} for {{Detection}} of
  {{Settlements Without Electricity}}.
\newblock In \emph{2021 {{IEEE International Geoscience}} and {{Remote Sensing
  Symposium}} (IGARSS)}, pages 1847--1850, 2021.
\newblock \doi{10.1109/IGARSS47720.2021.9554286}.

\bibitem[Diaconu et~al.(2022)Diaconu, Saha, Günnemann, and
  Zhu]{diaconu2022understandingroleweather}
Codruț-Andrei Diaconu, Sudipan Saha, Stephan Günnemann, and Xiao~Xiang Zhu.
\newblock Understanding the {{Role}} of {{Weather Data}} for {{Earth Surface
  Forecasting Using}} a {{ConvLSTM-Based Model}}.
\newblock In \emph{Proceedings of the IEEE/CVF Conference on Computer Vision
  and Pattern Recognition (CVPR)}, pages 1362--1371, 2022.

\bibitem[Doña et~al.(2015)Doña, Chang, Caselles, Sánchez, Camacho, Delegido,
  and Vannah]{dona2015integratedsatellitedata}
Carolina Doña, Ni-Bin Chang, Vicente Caselles, Juan~M. Sánchez, Antonio
  Camacho, Jesús Delegido, and Benjamin~W. Vannah.
\newblock Integrated satellite data fusion and mining for monitoring lake water
  quality status of the {{Albufera}} de {{Valencia}} in {{Spain}}.
\newblock \emph{Journal of Environmental Management}, 151:\penalty0 416--426,
  2015.
\newblock ISSN 0301-4797.
\newblock \doi{10.1016/j.jenvman.2014.12.003}.

\bibitem[Feichtenhofer et~al.(2016)Feichtenhofer, Pinz, and
  Zisserman]{feichtenhofer2016convolutionaltwostreamnetworka}
Christoph Feichtenhofer, Axel Pinz, and Andrew Zisserman.
\newblock Convolutional {{Two-Stream Network Fusion}} for {{Video Action
  Recognition}}.
\newblock In \emph{Proceedings of the IEEE/CVF Conference on Computer Vision
  and Pattern Recognition (CVPR)}, pages 1933--1941, 2016.

\bibitem[Gadiraju et~al.(2020)Gadiraju, Ramachandra, Chen, and
  Vatsavai]{gadiraju2020multimodaldeeplearninga}
Krishna~Karthik Gadiraju, Bharathkumar Ramachandra, Zexi Chen, and Ranga~Raju
  Vatsavai.
\newblock Multimodal {{Deep Learning Based Crop Classification Using
  Multispectral}} and {{Multitemporal Satellite Imagery}}.
\newblock In \emph{Proceedings of the {{International Conference}} on
  {{Knowledge Discovery}} \& {{Data Mining}} (SIGKDD)}, pages 3234--3242.
  {ACM}, 2020.
\newblock ISBN 978-1-4503-7998-4.
\newblock \doi{10.1145/3394486.3403375}.

\bibitem[Gangopadhyay et~al.()Gangopadhyay, Shook, Singh, and
  Sarkar]{gangopadhyay2019deeptimeseries}
Tryambak Gangopadhyay, Johnathon Shook, Asheesh~K Singh, and Soumik Sarkar.
\newblock Deep {{Time Series Attention Models}} for {{Crop Yield Prediction}}
  and {{Insights}}.
\newblock \emph{{{Neural Information Processing Systems}} (NIPS) Workshop on
  Machine Learning and the Physical Sciences}.

\bibitem[Gao et~al.(2006)Gao, Masek, Schwaller, and
  Hall]{gao2006blendinglandsatmodisa}
Feng Gao, J.~Masek, M.~Schwaller, and F.~Hall.
\newblock On the blending of the {{Landsat}} and {{MODIS}} surface reflectance:
  Predicting daily {{Landsat}} surface reflectance.
\newblock \emph{IEEE Transactions on Geoscience and Remote Sensing},
  44\penalty0 (8):\penalty0 2207--2218, 2006.
\newblock ISSN 1558-0644.
\newblock \doi{10.1109/TGRS.2006.872081}.

\bibitem[Gao et~al.(2022)Gao, Chen, and
  Xu]{gao2022adaptivespectralspatialfeature}
Hongmin Gao, Zhonghao Chen, and Feng Xu.
\newblock Adaptive spectral-spatial feature fusion network for hyperspectral
  image classification using limited training samples.
\newblock \emph{International Journal of Applied Earth Observation and
  Geoinformation}, 107:\penalty0 102687, 2022.
\newblock ISSN 0303-2434.
\newblock \doi{10.1016/j.jag.2022.102687}.

\bibitem[Gavahi et~al.(2021)Gavahi, Abbaszadeh, and
  Moradkhani]{gavahi2021deepyieldcombinedconvolutionala}
Keyhan Gavahi, Peyman Abbaszadeh, and Hamid Moradkhani.
\newblock {{DeepYield}}: {{A}} combined convolutional neural network with long
  short-term memory for crop yield forecasting.
\newblock \emph{Expert Systems with Applications}, 184:\penalty0 115511, 2021.
\newblock ISSN 0957-4174.
\newblock \doi{10.1016/j.eswa.2021.115511}.

\bibitem[Georgiou et~al.(2019)Georgiou, Papaioannou, and
  Potamianos]{georgiou2019deephierarchicalfusion}
Efthymios Georgiou, Charilaos Papaioannou, and Alexandros Potamianos.
\newblock Deep {{Hierarchical Fusion}} with {{Application}} in {{Sentiment
  Analysis}}.
\newblock In \emph{Interspeech 2019}, pages 1646--1650. {ISCA}, 2019.
\newblock \doi{10.21437/Interspeech.2019-3243}.

\bibitem[Ghamisi et~al.(2017)Ghamisi, Höfle, and
  Zhu]{ghamisi2017hyperspectrallidardata}
Pedram Ghamisi, Bernhard Höfle, and Xiao~Xiang Zhu.
\newblock Hyperspectral and {{LiDAR Data Fusion Using Extinction Profiles}} and
  {{Deep Convolutional Neural Network}}.
\newblock \emph{IEEE Journal of Selected Topics in Applied Earth Observations
  and Remote Sensing}, 10\penalty0 (6):\penalty0 3011--3024, 2017.
\newblock ISSN 2151-1535.
\newblock \doi{10.1109/JSTARS.2016.2634863}.

\bibitem[Ghamisi et~al.(2019)Ghamisi, Rasti, Yokoya, Wang, Hofle, Bruzzone,
  Bovolo, Chi, Anders, Gloaguen, Atkinson, and
  Benediktsson]{ghamisi2019multisourcemultitemporaldata}
Pedram Ghamisi, Behnood Rasti, Naoto Yokoya, Qunming Wang, Bernhard Hofle,
  Lorenzo Bruzzone, Francesca Bovolo, Mingmin Chi, Katharina Anders, Richard
  Gloaguen, Peter~M. Atkinson, and Jon~Atli Benediktsson.
\newblock Multisource and {{Multitemporal Data Fusion}} in {{Remote Sensing}}:
  {{A Comprehensive Review}} of the {{State}} of the {{Art}}.
\newblock \emph{IEEE Geoscience and Remote Sensing Magazine}, 7\penalty0
  (1):\penalty0 6--39, 2019.
\newblock ISSN 2168-6831.
\newblock \doi{10.1109/MGRS.2018.2890023}.

\bibitem[Gomez-Chova et~al.(2006)Gomez-Chova, Fernández-Prieto, Calpe, Soria,
  Vila, and Camps-Valls]{gomez-chova2006urbanmonitoringusing}
Luis Gomez-Chova, Diego Fernández-Prieto, Javier Calpe, Emilio Soria, Joan
  Vila, and Gustavo Camps-Valls.
\newblock Urban monitoring using multi-temporal {{SAR}} and multi-spectral
  data.
\newblock \emph{Pattern Recognition Letters}, 27\penalty0 (4):\penalty0
  234--243, 2006.
\newblock ISSN 0167-8655.
\newblock \doi{10.1016/j.patrec.2005.08.004}.

\bibitem[Gómez et~al.(2019)Gómez, Salvador, Sanz, and
  Casanova]{gomez2019potatoyieldpredictionb}
Diego Gómez, Pablo Salvador, Julia Sanz, and Jose~Luis Casanova.
\newblock Potato {{Yield Prediction Using Machine Learning Techniques}} and
  {{Sentinel}} 2 {{Data}}.
\newblock \emph{Remote Sensing}, 11\penalty0 (15):\penalty0 1745, 2019.
\newblock ISSN 2072-4292.
\newblock \doi{10.3390/rs11151745}.

\bibitem[Gómez-Chova et~al.(2015)Gómez-Chova, Tuia, Moser, and
  Camps-Valls]{gomez-chova2015multimodalclassificationremotea}
Luis Gómez-Chova, Devis Tuia, Gabriele Moser, and Gustau Camps-Valls.
\newblock Multimodal {{Classification}} of {{Remote Sensing Images}}: {{A
  Review}} and {{Future Directions}}.
\newblock \emph{Proceedings of the IEEE}, 103\penalty0 (9):\penalty0
  1560--1584, 2015.
\newblock ISSN 1558-2256.
\newblock \doi{10.1109/JPROC.2015.2449668}.

\bibitem[{H. Russello}(2018)]{h.russello2018convolutionalneuralnetworksa}
{H. Russello}.
\newblock Convolutional {{Neural Networks}} for {{Crop Yield Prediction}} using
  {{Satellite}}.pdf, 2018.

\bibitem[Hang et~al.(2020)Hang, Li, Ghamisi, Hong, Xia, and
  Liu]{hang2020classificationhyperspectrallidar}
Renlong Hang, Zhu Li, Pedram Ghamisi, Danfeng Hong, Guiyu Xia, and Qingshan
  Liu.
\newblock Classification of {{Hyperspectral}} and {{LiDAR Data Using Coupled
  CNNs}}.
\newblock \emph{IEEE Transactions on Geoscience and Remote Sensing},
  58\penalty0 (7):\penalty0 4939--4950, 2020.
\newblock ISSN 1558-0644.
\newblock \doi{10.1109/TGRS.2020.2969024}.

\bibitem[Hazirbas et~al.(2017)Hazirbas, Ma, Domokos, and
  Cremers]{hazirbas2017fusenetincorporatingdeptha}
Caner Hazirbas, Lingni Ma, Csaba Domokos, and Daniel Cremers.
\newblock {{FuseNet}}: {{Incorporating Depth}} into {{Semantic Segmentation}}
  via {{Fusion-Based CNN Architecture}}.
\newblock In \emph{Asian Conference on Computer Vision (ACCV)}, pages 213--228.
  {Springer International Publishing}, 2017.
\newblock ISBN 978-3-319-54181-5.
\newblock \doi{10.1007/978-3-319-54181-5_14}.

\bibitem[He et~al.(2016)He, Zhang, Ren, and Sun]{he2016deepresiduallearning}
Kaiming He, Xiangyu Zhang, Shaoqing Ren, and Jian Sun.
\newblock Deep {{Residual Learning}} for {{Image Recognition}}.
\newblock In \emph{Proceedings of the IEEE/CVF Conference on Computer Vision
  and Pattern Recognition (CVPR)}, pages 770--778, 2016.

\bibitem[Heidler et~al.(2021)Heidler, Mou, Hu, Jin, Li, Gan, Wen, and
  Zhu]{heidler2021selfsupervisedaudiovisualrepresentation}
Konrad Heidler, Lichao Mou, Di~Hu, Pu~Jin, Guangyao Li, Chuang Gan, Ji-Rong
  Wen, and Xiao~Xiang Zhu.
\newblock Self-supervised {{Audiovisual Representation Learning}} for {{Remote
  Sensing Data}}, 2021.

\bibitem[Hong et~al.(2021{\natexlab{a}})Hong, Gao, Yokoya, Yao, Chanussot, Du,
  and Zhang]{hong2021morediversemeansa}
Danfeng Hong, Lianru Gao, Naoto Yokoya, Jing Yao, Jocelyn Chanussot, Qian Du,
  and Bing Zhang.
\newblock More {{Diverse Means Better}}: {{Multimodal Deep Learning Meets
  Remote-Sensing Imagery Classification}}.
\newblock \emph{IEEE Transactions on Geoscience and Remote Sensing},
  59\penalty0 (5):\penalty0 4340--4354, 2021{\natexlab{a}}.
\newblock ISSN 1558-0644.
\newblock \doi{10.1109/TGRS.2020.3016820}.

\bibitem[Hong et~al.(2021{\natexlab{b}})Hong, Hu, Yao, Chanussot, and
  Zhu]{hong2021multimodalremotesensing}
Danfeng Hong, Jingliang Hu, Jing Yao, Jocelyn Chanussot, and Xiao~Xiang Zhu.
\newblock Multimodal remote sensing benchmark datasets for land cover
  classification with a shared and specific feature learning model.
\newblock \emph{ISPRS Journal of Photogrammetry and Remote Sensing},
  178:\penalty0 68--80, 2021{\natexlab{b}}.
\newblock ISSN 0924-2716.
\newblock \doi{10.1016/j.isprsjprs.2021.05.011}.

\bibitem[Hosseinpour et~al.(2022)Hosseinpour, Samadzadegan, and
  Javan]{hosseinpour2022cmgfnetdeepcrossmodala}
Hamidreza Hosseinpour, Farhad Samadzadegan, and Farzaneh~Dadrass Javan.
\newblock {{CMGFNet}}: {{A}} deep cross-modal gated fusion network for building
  extraction from very high-resolution remote sensing images.
\newblock \emph{ISPRS Journal of Photogrammetry and Remote Sensing},
  184:\penalty0 96--115, 2022.
\newblock ISSN 0924-2716.
\newblock \doi{10.1016/j.isprsjprs.2021.12.007}.

\bibitem[Hotelling(1936)]{hotelling1936relationstwosets}
Harold Hotelling.
\newblock Relations {{Between Two Sets}} of {{Variates}}.
\newblock \emph{Biometrika}, 28\penalty0 (3/4):\penalty0 321--377, 1936.
\newblock ISSN 0006-3444.
\newblock \doi{10.2307/2333955}.

\bibitem[Ienco et~al.(2019)Ienco, Interdonato, Gaetano, and
  Ho~Tong~Minh]{ienco2019combiningsentinel1sentinel2b}
Dino Ienco, Roberto Interdonato, Raffaele Gaetano, and Dinh Ho~Tong~Minh.
\newblock Combining {{Sentinel-1}} and {{Sentinel-2 Satellite Image Time
  Series}} for land cover mapping via a multi-source deep learning
  architecture.
\newblock \emph{ISPRS Journal of Photogrammetry and Remote Sensing},
  158:\penalty0 11--22, 2019.
\newblock ISSN 0924-2716.
\newblock \doi{10.1016/j.isprsjprs.2019.09.016}.

\bibitem[Ioffe and Szegedy(2015)]{ioffe2015batch}
Sergey Ioffe and Christian Szegedy.
\newblock Batch normalization: Accelerating deep network training by reducing
  internal covariate shift.
\newblock In \emph{Proceedings of International Conference on Machine Learning
  (ICML)}, pages 448--456. PMLR, 2015.

\bibitem[Irvin et~al.(2020)Irvin, Sheng, Ramachandran, Johnson-Yu, Zhou, Story,
  Rustowicz, Elsworth, Austin, and Ng]{irvin2020forestnetclassifyingdrivers}
Jeremy Irvin, Hao Sheng, Neel Ramachandran, Sonja Johnson-Yu, Sharon Zhou, Kyle
  Story, Rose Rustowicz, Cooper Elsworth, Kemen Austin, and Andrew~Y. Ng.
\newblock {{ForestNet}}: {{Classifying Drivers}} of {{Deforestation}} in
  {{Indonesia}} using {{Deep Learning}} on {{Satellite Imagery}}.
\newblock \emph{Advances in Neural Information Processing Systems (NIPS)}, 34,
  2020.

\bibitem[Jiang et~al.(2020)Jiang, Hu, Zhong, Xu, Xu, Huang, Wang, Ying, and
  Lin]{jiang2020deeplearningapproacha}
Hao Jiang, Hao Hu, Renhai Zhong, Jinfan Xu, Jialu Xu, Jingfeng Huang, Shaowen
  Wang, Yibin Ying, and Tao Lin.
\newblock A deep learning approach to conflating heterogeneous geospatial data
  for corn yield estimation: {{A}} case study of the {{US Corn Belt}} at the
  county level.
\newblock \emph{Global Change Biology}, 26\penalty0 (3):\penalty0 1754--1766,
  2020.
\newblock ISSN 1365-2486.
\newblock \doi{10.1111/gcb.14885}.

\bibitem[Jiang et~al.(2018)Jiang, Liu, Hendricks, Ganapathysubramanian, Hayes,
  and Sarkar]{jiang2018predictingcountylevel}
Zehui Jiang, Chao Liu, Nathan~P. Hendricks, Baskar Ganapathysubramanian,
  Dermot~J. Hayes, and Soumik Sarkar.
\newblock Predicting {{County Level Corn Yields Using Deep Long Short Term
  Memory Models}}, 2018.

\bibitem[Jiang(2019)]{jiang2019novelcropweed}
Zichao Jiang.
\newblock A {{Novel Crop Weed Recognition Method Based}} on {{Transfer
  Learning}} from {{VGG16 Implemented}} by {{Keras}}.
\newblock \emph{IOP Conference Series: Materials Science and Engineering},
  677\penalty0 (3):\penalty0 032073, 2019.
\newblock ISSN 1757-899X.
\newblock \doi{10.1088/1757-899X/677/3/032073}.

\bibitem[Johnson(2014)]{johnson2014assessmentprewithinseasona}
David~M. Johnson.
\newblock An assessment of pre- and within-season remotely sensed variables for
  forecasting corn and soybean yields in the {{United States}}.
\newblock \emph{Remote Sensing of Environment}, 141:\penalty0 116--128, 2014.
\newblock ISSN 0034-4257.
\newblock \doi{10.1016/j.rse.2013.10.027}.

\bibitem[Johnson et~al.(2016)Johnson, Hsieh, Cannon, Davidson, and
  Bédard]{johnson2016cropyieldforecasting}
Michael~D. Johnson, William~W. Hsieh, Alex~J. Cannon, Andrew Davidson, and
  Frédéric Bédard.
\newblock Crop yield forecasting on the {{Canadian Prairies}} by remotely
  sensed vegetation indices and machine learning methods.
\newblock \emph{Agricultural and Forest Meteorology}, 218--219:\penalty0
  74--84, 2016.
\newblock ISSN 0168-1923.
\newblock \doi{10.1016/j.agrformet.2015.11.003}.

\bibitem[Kang et~al.(2020)Kang, Ozdogan, Zhu, Ye, Hain, and
  Anderson]{kang2020comparativeassessmentenvironmentala}
Yanghui Kang, Mutlu Ozdogan, Xiaojin Zhu, Zhiwei Ye, Christopher Hain, and
  Martha Anderson.
\newblock Comparative assessment of environmental variables and machine
  learning algorithms for maize yield prediction in the {{US Midwest}}.
\newblock \emph{Environmental Research Letters}, 15\penalty0 (6):\penalty0
  064005, 2020.
\newblock ISSN 1748-9326.
\newblock \doi{10.1088/1748-9326/ab7df9}.

\bibitem[Kettenring(1971)]{kettenring1971canonicalanalysisseveral}
J.~R. Kettenring.
\newblock Canonical analysis of several sets of variables.
\newblock \emph{Biometrika}, 58\penalty0 (3):\penalty0 433--451, 1971.
\newblock ISSN 0006-3444.
\newblock \doi{10.1093/biomet/58.3.433}.

\bibitem[Khaki et~al.(2020)Khaki, Wang, and
  Archontoulis]{khaki2020cnnrnnframeworkcropa}
Saeed Khaki, Lizhi Wang, and Sotirios~V. Archontoulis.
\newblock A {{CNN-RNN Framework}} for {{Crop Yield Prediction}}.
\newblock \emph{Frontiers in Plant Science}, 10, 2020.
\newblock ISSN 1664-462X.
\newblock \doi{10.3389/fpls.2019.01750}.

\bibitem[Khaki et~al.(2021)Khaki, Pham, and
  Wang]{khaki2021simultaneouscornsoybeana}
Saeed Khaki, Hieu Pham, and Lizhi Wang.
\newblock Simultaneous corn and soybean yield prediction from remote sensing
  data using deep transfer learning.
\newblock \emph{Scientific Reports}, 11\penalty0 (1):\penalty0 11132, 2021.
\newblock ISSN 2045-2322.
\newblock \doi{10.1038/s41598-021-89779-z}.

\bibitem[Khodadadzadeh et~al.(2015)Khodadadzadeh, Li, Prasad, and
  Plaza]{khodadadzadeh2015fusionhyperspectrallidar}
Mahdi Khodadadzadeh, Jun Li, Saurabh Prasad, and Antonio Plaza.
\newblock Fusion of {{Hyperspectral}} and {{LiDAR Remote Sensing Data Using
  Multiple Feature Learning}}.
\newblock \emph{IEEE Journal of Selected Topics in Applied Earth Observations
  and Remote Sensing}, 8\penalty0 (6):\penalty0 2971--2983, 2015.
\newblock ISSN 2151-1535.
\newblock \doi{10.1109/JSTARS.2015.2432037}.

\bibitem[Kim and Lee(2016)]{kim2016machinelearningapproaches}
Nari Kim and Yang-Won Lee.
\newblock Machine {{Learning Approaches}} to {{Corn Yield Estimation Using
  Satellite Images}} and {{Climate Data}}: {{A Case}} of {{Iowa State}}.
\newblock \emph{Journal of the Korean Society of Surveying, Geodesy,
  Photogrammetry and Cartography}, 34\penalty0 (4):\penalty0 383--390, 2016.
\newblock ISSN 1598-4850.
\newblock \doi{10.7848/ksgpc.2016.34.4.383}.

\bibitem[Konapala et~al.(2021)Konapala, Kumar, and
  Khalique~Ahmad]{konapala2021exploringsentinel1sentinel2}
Goutam Konapala, Sujay~V. Kumar, and Shahryar Khalique~Ahmad.
\newblock Exploring {{Sentinel-1}} and {{Sentinel-2}} diversity for flood
  inundation mapping using deep learning.
\newblock \emph{ISPRS Journal of Photogrammetry and Remote Sensing},
  180:\penalty0 163--173, 2021.
\newblock ISSN 0924-2716.
\newblock \doi{10.1016/j.isprsjprs.2021.08.016}.

\bibitem[Kumar et~al.(2002)Kumar, Raghuwanshi, Singh, Wallender, and
  Pruitt]{kumar2002estimatingevapotranspirationusing}
M.~Kumar, N.~S. Raghuwanshi, R.~Singh, W.~W. Wallender, and W.~O. Pruitt.
\newblock Estimating {{Evapotranspiration}} using {{Artificial Neural
  Network}}.
\newblock \emph{Journal of Irrigation and Drainage Engineering}, 128\penalty0
  (4):\penalty0 224--233, 2002.
\newblock ISSN 0733-9437.
\newblock \doi{10.1061/(ASCE)0733-9437(2002)128:4(224)}.

\bibitem[Kumar et~al.(2008)Kumar, Bandyopadhyay, Raghuwanshi, and
  Singh]{kumar2008comparativestudyconventional}
M.~Kumar, A.~Bandyopadhyay, N.~S. Raghuwanshi, and R.~Singh.
\newblock Comparative study of conventional and artificial neural network-based
  {{ETo}} estimation models.
\newblock \emph{Irrigation Science}, 26\penalty0 (6):\penalty0 531, 2008.
\newblock ISSN 1432-1319.
\newblock \doi{10.1007/s00271-008-0114-3}.

\bibitem[Kussul et~al.(2017)Kussul, Lavreniuk, Skakun, and
  Shelestov]{kussul2017deeplearningclassificationa}
Nataliia Kussul, Mykola Lavreniuk, Sergii Skakun, and Andrii Shelestov.
\newblock Deep {{Learning Classification}} of {{Land Cover}} and {{Crop Types
  Using Remote Sensing Data}}.
\newblock \emph{IEEE Geoscience and Remote Sensing Letters}, 14\penalty0
  (5):\penalty0 778--782, 2017.
\newblock ISSN 1558-0571.
\newblock \doi{10.1109/LGRS.2017.2681128}.

\bibitem[Lahat et~al.(2015)Lahat, Adali, and
  Jutten]{lahat2015multimodaldatafusion}
Dana Lahat, Tülay Adali, and Christian Jutten.
\newblock Multimodal {{Data Fusion}}: {{An Overview}} of {{Methods}},
  {{Challenges}}, and {{Prospects}}.
\newblock \emph{Proceedings of the IEEE}, 103\penalty0 (9):\penalty0
  1449--1477, 2015.
\newblock ISSN 1558-2256.
\newblock \doi{10.1109/JPROC.2015.2460697}.

\bibitem[Lee et~al.(2015)Lee, Chan, Wilkin, and
  Remagnino]{lee2015deepplantplantidentification}
Sue~Han Lee, Chee~Seng Chan, Paul Wilkin, and Paolo Remagnino.
\newblock Deep-plant: {{Plant}} identification with convolutional neural
  networks.
\newblock In \emph{{{IEEE International Conference}} on {{Image Processing}}
  ({{ICIP}})}, pages 452--456, 2015.
\newblock \doi{10.1109/ICIP.2015.7350839}.

\bibitem[Lei et~al.(2022)Lei, Bai, Zhang, and
  Li]{lei2022convolutionneuralnetwork}
Dajiang Lei, Menghao Bai, Liping Zhang, and Weisheng Li.
\newblock Convolution neural network with edge structure loss for
  spatiotemporal remote sensing image fusion.
\newblock \emph{International Journal of Remote Sensing}, 43\penalty0
  (3):\penalty0 1015--1036, 2022.
\newblock ISSN 0143-1161.
\newblock \doi{10.1080/01431161.2022.2030070}.

\bibitem[Li et~al.(2022{\natexlab{a}})Li, Hong, Gao, Yao, Zheng, Zhang, and
  Chanussot]{li2022deeplearningmultimodal}
Jiaxin Li, Danfeng Hong, Lianru Gao, Jing Yao, Ke~Zheng, Bing Zhang, and
  Jocelyn Chanussot.
\newblock Deep learning in multimodal remote sensing data fusion: {{A}}
  comprehensive review.
\newblock \emph{International Journal of Applied Earth Observation and
  Geoinformation}, 112:\penalty0 102926, 2022{\natexlab{a}}.
\newblock ISSN 1569-8432.
\newblock \doi{10.1016/j.jag.2022.102926}.

\bibitem[Li et~al.(2019)Li, Yang, and
  Zhang]{li2019surveymultiviewrepresentation}
Yingming Li, Ming Yang, and Zhongfei Zhang.
\newblock A {{Survey}} of {{Multi-View Representation Learning}}.
\newblock \emph{IEEE Transactions on Knowledge and Data Engineering},
  31\penalty0 (10):\penalty0 1863--1883, 2019.
\newblock ISSN 1558-2191.
\newblock \doi{10.1109/TKDE.2018.2872063}.

\bibitem[Li et~al.(2022{\natexlab{b}})Li, Lu, Zhang, Tu, Li, Huang, Robinson,
  Malkin, Jojic, Ghamisi, Hänsch, and Yokoya]{li2022outcome2021ieee}
Zhuohong Li, Fangxiao Lu, Hongyan Zhang, Lilin Tu, Jiayi Li, Xin Huang, Caleb
  Robinson, Nikolay Malkin, Nebojsa Jojic, Pedram Ghamisi, Ronny Hänsch, and
  Naoto Yokoya.
\newblock The {{Outcome}} of the 2021 {{IEEE GRSS Data Fusion
  Contest}}—{{Track MSD}}: {{Multitemporal Semantic Change Detection}}.
\newblock \emph{IEEE Journal of Selected Topics in Applied Earth Observations
  and Remote Sensing}, 15:\penalty0 1643--1655, 2022{\natexlab{b}}.
\newblock ISSN 2151-1535.
\newblock \doi{10.1109/JSTARS.2022.3144318}.

\bibitem[Lin et~al.(2020)Lin, Zhong, Wang, Xu, Jiang, Xu, Ying, Rodriguez,
  Ting, and Li]{lin2020deepcropnetdeepspatialtemporal}
Tao Lin, Renhai Zhong, Yudi Wang, Jinfan Xu, Hao Jiang, Jialu Xu, Yibin Ying,
  Luis Rodriguez, K.~C. Ting, and Haifeng Li.
\newblock {{DeepCropNet}}: A deep spatial-temporal learning framework for
  county-level corn yield estimation.
\newblock \emph{Environmental Research Letters}, 15\penalty0 (3):\penalty0
  034016, 2020.
\newblock ISSN 1748-9326.
\newblock \doi{10.1088/1748-9326/ab66cb}.

\bibitem[Liu et~al.(2020)Liu, Kampffmeyer, Jenssen, and
  Salberg]{liu2020multiviewselfconstructinggrapha}
Qinghui Liu, Michael~C. Kampffmeyer, Robert Jenssen, and Arnt-Borre Salberg.
\newblock Multi-{{View Self-Constructing Graph Convolutional Networks With
  Adaptive Class Weighting Loss}} for {{Semantic Segmentation}}.
\newblock In \emph{Proceedings of the IEEE/CVF Conference on Computer Vision
  and Pattern Recognition (CVPR) Workshops}, pages 44--45, 2020.

\bibitem[Liu and Abd-Elrahman(2018)]{liu2018deepconvolutionalneural}
Tao Liu and Amr Abd-Elrahman.
\newblock Deep convolutional neural network training enrichment using
  multi-view object-based analysis of unmanned aerial systems imagery for
  wetlands classification.
\newblock \emph{ISPRS Journal of Photogrammetry and Remote Sensing},
  139:\penalty0 154--170, 2018.
\newblock \doi{10.1016/j.isprsjprs.2018.03.006}.

\bibitem[Livieris et~al.(2020)Livieris, Dafnis, Papadopoulos, and
  Kalivas]{livieris2020multipleinputneuralnetworka}
Ioannis~E. Livieris, Spiros~D. Dafnis, George~K. Papadopoulos, and
  Dionissios~P. Kalivas.
\newblock A {{Multiple-Input Neural Network Model}} for {{Predicting Cotton
  Production Quantity}}: {{A Case Study}}.
\newblock \emph{Algorithms}, 13\penalty0 (11):\penalty0 273, 2020.
\newblock ISSN 1999-4893.
\newblock \doi{10.3390/a13110273}.

\bibitem[Luus et~al.(2015)Luus, Salmon, van~den Bergh, and
  Maharaj]{luus2015multiviewdeeplearning}
F.~P.~S. Luus, B.~P. Salmon, F.~van~den Bergh, and B.~T.~J. Maharaj.
\newblock Multiview {{Deep Learning}} for {{Land-Use Classification}}.
\newblock \emph{IEEE Geoscience and Remote Sensing Letters}, 12\penalty0
  (12):\penalty0 2448--2452, 2015.
\newblock ISSN 1558-0571.
\newblock \doi{10.1109/LGRS.2015.2483680}.

\bibitem[Ma et~al.(2021)Ma, Li, Feng, Xia, Huang, Zhang, Prieur, Licciardi,
  Malha, Chanussot, Ghamisi, Hänsch, and Yokoya]{ma2021outcome2021ieee}
Yanbiao Ma, Yuxin Li, Kexin Feng, Yu~Xia, Qi~Huang, Hongyan Zhang, Colin
  Prieur, Giorgio Licciardi, Hana Malha, Jocelyn Chanussot, Pedram Ghamisi,
  Ronny Hänsch, and Naoto Yokoya.
\newblock The {{Outcome}} of the 2021 {{IEEE GRSS Data Fusion Contest}} -
  {{Track DSE}}: {{Detection}} of {{Settlements Without Electricity}}.
\newblock \emph{IEEE Journal of Selected Topics in Applied Earth Observations
  and Remote Sensing}, 14:\penalty0 12375--12385, 2021.
\newblock ISSN 2151-1535.
\newblock \doi{10.1109/JSTARS.2021.3130446}.

\bibitem[Maimaitijiang et~al.(2020)Maimaitijiang, Sagan, Sidike, Hartling,
  Esposito, and Fritschi]{maimaitijiang2020soybeanyieldpredictiona}
Maitiniyazi Maimaitijiang, Vasit Sagan, Paheding Sidike, Sean Hartling, Flavio
  Esposito, and Felix~B. Fritschi.
\newblock Soybean yield prediction from {{UAV}} using multimodal data fusion
  and deep learning.
\newblock \emph{Remote Sensing of Environment}, 237:\penalty0 111599, 2020.
\newblock ISSN 0034-4257.
\newblock \doi{10.1016/j.rse.2019.111599}.

\bibitem[Mantsis et~al.(2022)Mantsis, Bakratsas, Andreadis, Karsisto,
  Moumtzidou, Gialampoukidis, Karppinen, Vrochidis, and
  Kompatsiaris]{mantsis2022multimodalfusionsentinel}
Damianos~Florin Mantsis, Marios Bakratsas, Stelios Andreadis, Petteri Karsisto,
  Anastasia Moumtzidou, Ilias Gialampoukidis, Ari Karppinen, Stefanos
  Vrochidis, and Ioannis Kompatsiaris.
\newblock Multimodal {{Fusion}} of {{Sentinel}} 1 {{Images}} and {{Social Media
  Data}} for {{Snow Depth Estimation}}.
\newblock \emph{IEEE Geoscience and Remote Sensing Letters}, 19:\penalty0 1--5,
  2022.
\newblock ISSN 1558-0571.
\newblock \doi{10.1109/LGRS.2020.3031866}.

\bibitem[Mao et~al.(2018)Mao, Yuan, and
  Xiaoqiang]{mao2018deepcrossmodalretrieval}
Guo Mao, Yuan Yuan, and Lu~Xiaoqiang.
\newblock Deep {{Cross-Modal Retrieval}} for {{Remote Sensing Image}} and
  {{Audio}}.
\newblock In \emph{{{IAPR Workshop}} on {{Pattern Recognition}} in {{Remote
  Sensing}} ({{PRRS}})}, pages 1--7, 2018.
\newblock \doi{10.1109/PRRS.2018.8486338}.

\bibitem[Marmanis et~al.(2016)Marmanis, Wegner, Galliani, Schindler, Datcu, and
  Stilla]{marmanis2016semanticsegmentationaerial}
Dimitrios Marmanis, Jan~D. Wegner, Silvano Galliani, K.~Schindler, Mihai Datcu,
  and U.~Stilla.
\newblock Semantic segmentation of aerial images with an ensemble of cnss.
\newblock In \emph{ISPRS Annals of the Photogrammetry, Remote Sensing and
  Spatial Information Sciences, 2016}, volume III-3, pages 473--480.
  {Copernicus Publications}, 2016.
\newblock \doi{10.5194/isprsannals-III-3-473-2016}.

\bibitem[Meng et~al.(2021)Meng, Liu, L.~Ustin, and
  Zhang]{meng2021predictingmaizeyielda}
Linghua Meng, Huanjun Liu, Susan L.~Ustin, and Xinle Zhang.
\newblock Predicting {{Maize Yield}} at the {{Plot Scale}} of {{Different
  Fertilizer Systems}} by {{Multi-Source Data}} and {{Machine Learning
  Methods}}.
\newblock \emph{Remote Sensing}, 13\penalty0 (18):\penalty0 3760, 2021.
\newblock ISSN 2072-4292.
\newblock \doi{10.3390/rs13183760}.

\bibitem[Mohanty et~al.(2016)Mohanty, Hughes, and
  Salathé]{mohanty2016usingdeeplearning}
Sharada~P. Mohanty, David~P. Hughes, and Marcel Salathé.
\newblock Using {{Deep Learning}} for {{Image-Based Plant Disease Detection}}.
\newblock \emph{Frontiers in Plant Science}, 7, 2016.
\newblock ISSN 1664-462X.

\bibitem[Mohla et~al.(2020)Mohla, Pande, Banerjee, and
  Chaudhuri]{mohla2020fusatnetdualattention}
Satyam Mohla, Shivam Pande, Biplab Banerjee, and Subhasis Chaudhuri.
\newblock {{FusAtNet}}: {{Dual Attention}} based {{SpectroSpatial Multimodal
  Fusion Network}} for {{Hyperspectral}} and {{LiDAR Classification}}.
\newblock In \emph{Proceedings of {{IEEE}}/{{CVF Conference}} on {{Computer
  Vision}} and {{Pattern Recognition}} ({{CVPR}}) Workshop}, pages 416--425,
  2020.
\newblock \doi{10.1109/CVPRW50498.2020.00054}.

\bibitem[Muslea et~al.(2002)Muslea, Minton, and
  Knoblock]{muslea2002activesemisupervisedlearning}
Ion Muslea, Steven Minton, and Craig~A. Knoblock.
\newblock Active + {{Semi-supervised Learning}} = {{Robust Multi-View
  Learning}}.
\newblock In \emph{Proceedings of {{International Conference}} on {{Machine
  Learning}} (ICML)}, pages 435--442. {Morgan Kaufmann Publishers Inc.}, 2002.
\newblock ISBN 978-1-55860-873-3.

\bibitem[Nemmour and Chibani(2006)]{nemmour2006multiplesupportvector}
Hassiba Nemmour and Youcef Chibani.
\newblock Multiple support vector machines for land cover change detection:
  {{An}} application for mapping urban extensions.
\newblock \emph{ISPRS Journal of Photogrammetry and Remote Sensing},
  61\penalty0 (2):\penalty0 125--133, 2006.
\newblock ISSN 0924-2716.
\newblock \doi{10.1016/j.isprsjprs.2006.09.004}.

\bibitem[Nevavuori et~al.(2019)Nevavuori, Narra, and
  Lipping]{nevavuori2019cropyieldpredictiona}
Petteri Nevavuori, Nathaniel Narra, and Tarmo Lipping.
\newblock Crop yield prediction with deep convolutional neural networks.
\newblock \emph{Computers and Electronics in Agriculture}, 163:\penalty0
  104859, 2019.
\newblock ISSN 0168-1699.
\newblock \doi{10.1016/j.compag.2019.104859}.

\bibitem[Nguyen et~al.(2019)Nguyen, Zhu, Lin, Du, Yang, Guo, and
  Jin]{nguyen2019spatialtemporalmultitasklearninga}
Long~H. Nguyen, Jiazhen Zhu, Zhe Lin, Hanxiang Du, Zhou Yang, Wenxuan Guo, and
  Fang Jin.
\newblock Spatial-{{Temporal Multi-Task Learning}} for {{Within-Field Cotton
  Yield Prediction}}.
\newblock In \emph{Advances in {{Knowledge Discovery}} and {{Data Mining}}},
  Lecture {{Notes}} in {{Computer Science}}, pages 343--354. {Springer
  International Publishing}, 2019.
\newblock ISBN 978-3-030-16148-4.
\newblock \doi{10.1007/978-3-030-16148-4_27}.

\bibitem[Niedbała(2019)]{niedbala2019simplemodelbaseda}
Gniewko Niedbała.
\newblock Simple model based on artificial neural network for early prediction
  and simulation winter rapeseed yield.
\newblock \emph{Journal of Integrative Agriculture}, 18\penalty0 (1):\penalty0
  54--61, 2019.
\newblock ISSN 2095-3119.
\newblock \doi{10.1016/S2095-3119(18)62110-0}.

\bibitem[Nijhawan et~al.()Nijhawan, Sharma, Sahni, and
  Batra]{nijhawan2017deeplearninghybrid}
Rahul Nijhawan, Himanshu Sharma, Harshita Sahni, and Ashita Batra.
\newblock A {{Deep Learning Hybrid CNN Framework Approach}} for {{Vegetation
  Cover Mapping Using Deep Features}}.
\newblock In \emph{Proceedings of {{International Conference}} on
  {{Signal-Image Technology Internet-Based Systems}} ({{SITIS}})}, pages
  192--196.
\newblock \doi{10.1109/SITIS.2017.41}.

\bibitem[Nowakowski et~al.(2021)Nowakowski, Mrziglod, Spiller, Bonifacio,
  Ferrari, Mathieu, Garcia-Herranz, and Kim]{nowakowski2021croptypemappinga}
Artur Nowakowski, John Mrziglod, Dario Spiller, Rogerio Bonifacio, Irene
  Ferrari, Pierre~Philippe Mathieu, Manuel Garcia-Herranz, and Do-Hyung Kim.
\newblock Crop type mapping by using transfer learning.
\newblock \emph{International Journal of Applied Earth Observation and
  Geoinformation}, 98:\penalty0 102313, 2021.
\newblock ISSN 0303-2434.
\newblock \doi{10.1016/j.jag.2021.102313}.

\bibitem[Ofori-Ampofo et~al.(2021)Ofori-Ampofo, Pelletier, and
  Lang]{ofori-ampofo2021croptypemappingb}
Stella Ofori-Ampofo, Charlotte Pelletier, and Stefan Lang.
\newblock Crop {{Type Mapping}} from {{Optical}} and {{Radar Time Series Using
  Attention-Based Deep Learning}}.
\newblock \emph{Remote Sensing}, 13\penalty0 (22):\penalty0 4668, 2021.
\newblock ISSN 2072-4292.
\newblock \doi{10.3390/rs13224668}.

\bibitem[Pageot et~al.(2020)Pageot, Baup, Inglada, Baghdadi, and
  Demarez]{pageot2020detectionirrigatedrainfed}
Yann Pageot, Frédéric Baup, Jordi Inglada, Nicolas Baghdadi, and Valérie
  Demarez.
\newblock Detection of {{Irrigated}} and {{Rainfed Crops}} in {{Temperate Areas
  Using Sentinel-1}} and {{Sentinel-2 Time Series}}.
\newblock \emph{Remote Sensing}, 12\penalty0 (18):\penalty0 3044, 2020.
\newblock ISSN 2072-4292.
\newblock \doi{10.3390/rs12183044}.

\bibitem[Pei et~al.(2018)Pei, Huang, Huo, Zhang, Yang, and
  Yeo]{pei2018sarautomatictarget}
Jifang Pei, Yulin Huang, Weibo Huo, Yin Zhang, Jianyu Yang, and Tat-Soon Yeo.
\newblock {{SAR Automatic Target Recognition Based}} on {{Multiview Deep
  Learning Framework}}.
\newblock \emph{IEEE Transactions on Geoscience and Remote Sensing},
  56\penalty0 (4):\penalty0 2196--2210, 2018.
\newblock ISSN 1558-0644.
\newblock \doi{10.1109/TGRS.2017.2776357}.

\bibitem[Peng et~al.(2020)Peng, Guan, Zhou, Jiang, Frankenberg, Sun, He, and
  Köhler]{peng2020assessingbenefitsatellitebaseda}
Bin Peng, Kaiyu Guan, Wang Zhou, Chongya Jiang, Christian Frankenberg, Ying
  Sun, Liyin He, and Philipp Köhler.
\newblock Assessing the benefit of satellite-based {{Solar-Induced Chlorophyll
  Fluorescence}} in crop yield prediction.
\newblock \emph{International Journal of Applied Earth Observation and
  Geoinformation}, 90:\penalty0 102126, 2020.
\newblock ISSN 0303-2434.
\newblock \doi{10.1016/j.jag.2020.102126}.

\bibitem[Perez-Rua et~al.(2019)Perez-Rua, Vielzeuf, Pateux, Baccouche, and
  Jurie]{perez-rua2019mfasmultimodalfusion}
Juan-Manuel Perez-Rua, Valentin Vielzeuf, Stephane Pateux, Moez Baccouche, and
  Frederic Jurie.
\newblock {{MFAS}}: {{Multimodal Fusion Architecture Search}}.
\newblock In \emph{Proceedings of the IEEE/CVF Conference on Computer Vision
  and Pattern Recognition}, pages 6966--6975, 2019.

\bibitem[Ramachandram and Taylor(2017)]{ramachandram2017deepmultimodallearning}
Dhanesh Ramachandram and Graham~W. Taylor.
\newblock Deep {{Multimodal Learning}}: {{A Survey}} on {{Recent Advances}} and
  {{Trends}}.
\newblock \emph{IEEE Signal Processing Magazine}, 34\penalty0 (6):\penalty0
  96--108, 2017.
\newblock ISSN 1558-0792.
\newblock \doi{10.1109/MSP.2017.2738401}.

\bibitem[Rambour et~al.(2020)Rambour, Audebert, Koeniguer, Le~Saux, Crucianu,
  and Datcu]{rambour2020flooddetectiontime}
C.~Rambour, N.~Audebert, E.~Koeniguer, B.~Le~Saux, M.~Crucianu, and M.~Datcu.
\newblock {{FLOOD DETECTION IN TIME SERIES OF OPTICAL AND SAR IMAGES}}.
\newblock In \emph{The {{International Archives}} of the {{Photogrammetry}},
  {{Remote Sensing}} and {{Spatial Information Sciences}}}, volume
  XLIII-B2-2020, pages 1343--1346. {Copernicus GmbH}, 2020.
\newblock \doi{10.5194/isprs-archives-XLIII-B2-2020-1343-2020}.

\bibitem[Ramcharan et~al.(2017)Ramcharan, Baranowski, McCloskey, Ahmed, Legg,
  and Hughes]{ramcharan2017deeplearningimagebased}
Amanda Ramcharan, Kelsee Baranowski, Peter McCloskey, Babuali Ahmed, James
  Legg, and David~P. Hughes.
\newblock Deep {{Learning}} for {{Image-Based Cassava Disease Detection}}.
\newblock \emph{Frontiers in Plant Science}, 8, 2017.
\newblock ISSN 1664-462X.

\bibitem[Rashkovetsky et~al.(2021)Rashkovetsky, Mauracher, Langer, and
  Schmitt]{rashkovetsky2021wildfiredetectionmultisensor}
Dmitry Rashkovetsky, Florian Mauracher, Martin Langer, and Michael Schmitt.
\newblock Wildfire {{Detection From Multisensor Satellite Imagery Using Deep
  Semantic Segmentation}}.
\newblock \emph{IEEE Journal of Selected Topics in Applied Earth Observations
  and Remote Sensing}, 14:\penalty0 7001--7016, 2021.
\newblock ISSN 2151-1535.
\newblock \doi{10.1109/JSTARS.2021.3093625}.

\bibitem[Reyes et~al.(2015)Reyes, Caicedo, and
  Camargo]{reyes2015finetuningdeepconvolutional}
Angie~K. Reyes, Juan~C. Caicedo, and Jorge~E. Camargo.
\newblock Fine-tuning {{Deep Convolutional Networks}} for {{Plant
  Recognition}}.
\newblock \emph{Proceedings of {{Conference}} and {{Labs}} of the
  {{Evaluation}} Forum (CLEF)}, 1391, 2015.

\bibitem[Robinson et~al.(2021)Robinson, Malkin, Jojic, Chen, Qin, Xiao,
  Schmitt, Ghamisi, Hänsch, and Yokoya]{robinson2021globallandcovermapping}
Caleb Robinson, Kolya Malkin, Nebojsa Jojic, Huijun Chen, Rongjun Qin, Changlin
  Xiao, Michael Schmitt, Pedram Ghamisi, Ronny Hänsch, and Naoto Yokoya.
\newblock Global {{Land-Cover Mapping With Weak Supervision}}: {{Outcome}} of
  the 2020 {{IEEE GRSS Data Fusion Contest}}.
\newblock \emph{IEEE Journal of Selected Topics in Applied Earth Observations
  and Remote Sensing}, 14:\penalty0 3185--3199, 2021.
\newblock ISSN 2151-1535.
\newblock \doi{10.1109/JSTARS.2021.3063849}.

\bibitem[Rudner et~al.(2019)Rudner, Rußwurm, Fil, Pelich, Bischke,
  Kopačková, and Biliński]{rudner2019multi3netsegmentingflooded}
Tim G.~J. Rudner, Marc Rußwurm, Jakub Fil, Ramona Pelich, Benjamin Bischke,
  Veronika Kopačková, and Piotr Biliński.
\newblock {{Multi3Net}}: {{Segmenting Flooded Buildings}} via {{Fusion}} of
  {{Multiresolution}}, {{Multisensor}}, and {{Multitemporal Satellite
  Imagery}}.
\newblock \emph{Proceedings of the AAAI Conference on Artificial Intelligence},
  33\penalty0 (01):\penalty0 702--709, 2019.
\newblock ISSN 2374-3468.
\newblock \doi{10.1609/aaai.v33i01.3301702}.

\bibitem[Ruß et~al.(2008)Ruß, Kruse, Schneider, and
  Wagner]{russ2008dataminingneural}
Georg Ruß, Rudolf Kruse, Martin Schneider, and Peter Wagner.
\newblock Data {{Mining}} with {{Neural Networks}} for {{Wheat Yield
  Prediction}}.
\newblock In \emph{Advances in {{Data Mining}}. {{Medical Applications}},
  {{E-Commerce}}, {{Marketing}}, and {{Theoretical Aspects}}}, Lecture
  {{Notes}} in {{Computer Science}}, pages 47--56. {Springer}, 2008.
\newblock ISBN 978-3-540-70720-2.
\newblock \doi{10.1007/978-3-540-70720-2_4}.

\bibitem[Sa et~al.(2016)Sa, Ge, Dayoub, Upcroft, Perez, and
  McCool]{sa2016deepfruitsfruitdetection}
Inkyu Sa, Zongyuan Ge, Feras Dayoub, Ben Upcroft, Tristan Perez, and Chris
  McCool.
\newblock {{DeepFruits}}: {{A Fruit Detection System Using Deep Neural
  Networks}}.
\newblock \emph{Sensors}, 16\penalty0 (8):\penalty0 1222, 2016.
\newblock ISSN 1424-8220.
\newblock \doi{10.3390/s16081222}.

\bibitem[Sagan et~al.(2021)Sagan, Maimaitijiang, Bhadra, Maimaitiyiming, Brown,
  Sidike, and Fritschi]{sagan2021fieldscalecropyielda}
Vasit Sagan, Maitiniyazi Maimaitijiang, Sourav Bhadra, Matthew Maimaitiyiming,
  Davis~R. Brown, Paheding Sidike, and Felix~B. Fritschi.
\newblock Field-scale crop yield prediction using multi-temporal
  {{WorldView-3}} and {{PlanetScope}} satellite data and deep learning.
\newblock \emph{ISPRS Journal of Photogrammetry and Remote Sensing},
  174:\penalty0 265--281, 2021.
\newblock ISSN 0924-2716.
\newblock \doi{10.1016/j.isprsjprs.2021.02.008}.

\bibitem[Sahu and Vechtomova(2021)]{sahu2021adaptivefusiontechniques}
Gaurav Sahu and Olga Vechtomova.
\newblock Adaptive {{Fusion Techniques}} for {{Multimodal Data}}.
\newblock In \emph{Proceedings of the {{Conference}} of the {{European
  Chapter}} of the {{Association}} for {{Computational Linguistics}}}, pages
  3156--3166. {Association for Computational Linguistics}, 2021.
\newblock \doi{10.18653/v1/2021.eacl-main.275}.

\bibitem[Said et~al.(2019)Said, Ahmad, Riegler, Pogorelov, Hassan, Ahmad, and
  Conci]{said2019naturaldisastersdetection}
Naina Said, Kashif Ahmad, Michael Riegler, Konstantin Pogorelov, Laiq Hassan,
  Nasir Ahmad, and Nicola Conci.
\newblock Natural disasters detection in social media and satellite imagery: a
  survey.
\newblock \emph{Multimedia Tools and Applications}, 78\penalty0 (22):\penalty0
  31267--31302, 2019.
\newblock \doi{10.1007/s11042-019-07942-1}.

\bibitem[Sainte Fare~Garnot et~al.(2020)Sainte Fare~Garnot, Landrieu, and
  Chehata]{saintefaregarnot2022multimodaltemporalattention}
Vivien Sainte Fare~Garnot, Loic Landrieu, and Nesrine Chehata.
\newblock Multi-modal temporal attention models for crop mapping from satellite
  time series.
\newblock \emph{ISPRS Journal of Photogrammetry and Remote Sensing},
  187:\penalty0 294--305, 2020.
\newblock ISSN 0924-2716.
\newblock \doi{10.1016/j.isprsjprs.2022.03.012}.

\bibitem[Salcedo-Sanz et~al.(2020)Salcedo-Sanz, Ghamisi, Piles, Werner, Cuadra,
  Moreno-Martínez, Izquierdo-Verdiguier, Muñoz-Marí, Mosavi, and
  Camps-Valls]{salcedo-sanz2020machinelearninginformationa}
S.~Salcedo-Sanz, P.~Ghamisi, M.~Piles, M.~Werner, L.~Cuadra,
  A.~Moreno-Martínez, E.~Izquierdo-Verdiguier, J.~Muñoz-Marí, Amirhosein
  Mosavi, and G.~Camps-Valls.
\newblock Machine learning information fusion in {{Earth}} observation: {{A}}
  comprehensive review of methods, applications and data sources.
\newblock \emph{Information Fusion}, 63:\penalty0 256--272, 2020.
\newblock ISSN 1566-2535.
\newblock \doi{10.1016/j.inffus.2020.07.004}.

\bibitem[Scarpa et~al.(2018)Scarpa, Gargiulo, Mazza, and
  Gaetano]{scarpa2018cnnbasedfusionmethod}
Giuseppe Scarpa, Massimiliano Gargiulo, Antonio Mazza, and Raffaele Gaetano.
\newblock A {{CNN-Based Fusion Method}} for {{Feature Extraction}} from
  {{Sentinel Data}}.
\newblock \emph{Remote Sensing}, 10\penalty0 (2):\penalty0 236, 2018.
\newblock ISSN 2072-4292.
\newblock \doi{10.3390/rs10020236}.

\bibitem[Scott et~al.(2018)Scott, Hagan, Marcum, Hurt, Anderson, and
  Davis]{scott2018enhancedfusiondeep}
Grant~J. Scott, Kyle~C. Hagan, Richard~A. Marcum, James~Alex Hurt, Derek~T.
  Anderson, and Curt~H. Davis.
\newblock Enhanced {{Fusion}} of {{Deep Neural Networks}} for
  {{Classification}} of {{Benchmark High-Resolution Image Data Sets}}.
\newblock \emph{IEEE Geoscience and Remote Sensing Letters}, 15\penalty0
  (9):\penalty0 1451--1455, 2018.
\newblock ISSN 1558-0571.
\newblock \doi{10.1109/LGRS.2018.2839092}.

\bibitem[Sebastianelli et~al.(2021)Sebastianelli, Del~Rosso, Mathieu, and
  Ullo]{sebastianelli2021paradigmselectiondata}
Alessandro Sebastianelli, Maria~Pia Del~Rosso, Pierre~Philippe Mathieu, and
  Silvia~Liberata Ullo.
\newblock Paradigm selection for {{Data Fusion}} of {{SAR}} and {{Multispectral
  Sentinel}} data applied to {{Land-Cover Classification}}, 2021.

\bibitem[Shahhosseini et~al.(2021)Shahhosseini, Hu, Khaki, and
  Archontoulis]{shahhosseini2021cornyieldpredictiona}
Mohsen Shahhosseini, Guiping Hu, Saeed Khaki, and Sotirios~V. Archontoulis.
\newblock Corn {{Yield Prediction With Ensemble CNN-DNN}}.
\newblock \emph{Frontiers in Plant Science}, 12:\penalty0 709008, 2021.
\newblock ISSN 1664-462X.
\newblock \doi{10.3389/fpls.2021.709008}.

\bibitem[Shen et~al.(2015)Shen, Li, Cheng, Zeng, Yang, Li, and
  Zhang]{shen2015missinginformationreconstruction}
Huanfeng Shen, Xinghua Li, Qing Cheng, Chao Zeng, Gang Yang, Huifang Li, and
  Liangpei Zhang.
\newblock Missing {{Information Reconstruction}} of {{Remote Sensing Data}}:
  {{A Technical Review}}.
\newblock \emph{IEEE Geoscience and Remote Sensing Magazine}, 3\penalty0
  (3):\penalty0 61--85, 2015.
\newblock ISSN 2168-6831.
\newblock \doi{10.1109/MGRS.2015.2441912}.

\bibitem[Shen et~al.(2016)Shen, Meng, and
  Zhang]{shen2016integratedframeworkspatioa}
Huanfeng Shen, Xiangchao Meng, and Liangpei Zhang.
\newblock An {{Integrated Framework}} for the
  {{Spatio}}–{{Temporal}}–{{Spectral Fusion}} of {{Remote Sensing Images}}.
\newblock \emph{IEEE Transactions on Geoscience and Remote Sensing},
  54\penalty0 (12):\penalty0 7135--7148, 2016.
\newblock ISSN 1558-0644.
\newblock \doi{10.1109/TGRS.2016.2596290}.

\bibitem[Song et~al.(2021)Song, Huang, Hansen, and
  Potapov]{song2021evaluationlandsatsentinel2}
Xiao-Peng Song, Wenli Huang, Matthew~C. Hansen, and Peter Potapov.
\newblock An evaluation of {{Landsat}}, {{Sentinel-2}}, {{Sentinel-1}} and
  {{MODIS}} data for crop type mapping.
\newblock \emph{Science of Remote Sensing}, 3:\penalty0 100018, 2021.
\newblock ISSN 2666-0172.
\newblock \doi{10.1016/j.srs.2021.100018}.

\bibitem[Srivastava et~al.(2014)Srivastava, Hinton, Krizhevsky, Sutskever, and
  Salakhutdinov]{srivastava2014dropout}
Nitish Srivastava, Geoffrey Hinton, Alex Krizhevsky, Ilya Sutskever, and Ruslan
  Salakhutdinov.
\newblock Dropout: a simple way to prevent neural networks from overfitting.
\newblock \emph{Journal of Machine Learning Research (JMLR)}, 15\penalty0
  (1):\penalty0 1929--1958, 2014.

\bibitem[Srivastava et~al.(2019)Srivastava, Vargas-Muñoz, and
  Tuia]{srivastava2019understandingurbanlanduse}
Shivangi Srivastava, John~E. Vargas-Muñoz, and Devis Tuia.
\newblock Understanding urban landuse from the above and ground perspectives:
  {{A}} deep learning, multimodal solution.
\newblock \emph{Remote Sensing of Environment}, 228:\penalty0 129--143, 2019.
\newblock ISSN 0034-4257.
\newblock \doi{10.1016/j.rse.2019.04.014}.

\bibitem[Srivastava et~al.(2020)Srivastava, Vargas~Muñoz, Lobry, and
  Tuia]{srivastava2020finegrainedlandusecharacterization}
Shivangi Srivastava, John~E. Vargas~Muñoz, Sylvain Lobry, and Devis Tuia.
\newblock Fine-grained landuse characterization using ground-based pictures: A
  deep learning solution based on globally available data.
\newblock \emph{International Journal of Geographical Information Science},
  34\penalty0 (6):\penalty0 1117--1136, 2020.
\newblock ISSN 1365-8816.
\newblock \doi{10.1080/13658816.2018.1542698}.

\bibitem[Stojnic and Risojevic(2021)]{stojnic2021selfsupervisedlearningremote}
Vladan Stojnic and Vladimir Risojevic.
\newblock Self-{{Supervised Learning}} of {{Remote Sensing Scene
  Representations Using Contrastive Multiview Coding}}.
\newblock In \emph{Proceedings of the IEEE/CVF Conference on Computer Vision
  and Pattern Recognition (CVPR)}, pages 1182--1191, 2021.

\bibitem[Summaira et~al.(2021)Summaira, Li, Shoib, Li, and
  Abdul]{summaira2021recentadvancestrends}
Jabeen Summaira, Xi~Li, Amin~Muhammad Shoib, Songyuan Li, and Jabbar Abdul.
\newblock Recent {{Advances}} and {{Trends}} in {{Multimodal Deep Learning}}:
  {{A Review}}, 2021.

\bibitem[Sun et~al.(2019)Sun, Di, Sun, Shen, and
  Lai]{sun2019countylevelsoybeanyielda}
Jie Sun, Liping Di, Ziheng Sun, Yonglin Shen, and Zulong Lai.
\newblock County-{{Level Soybean Yield Prediction Using Deep CNN-LSTM Model}}.
\newblock \emph{Sensors}, 19\penalty0 (20):\penalty0 4363, 2019.
\newblock ISSN 1424-8220.
\newblock \doi{10.3390/s19204363}.

\bibitem[Sun(2013)]{sun2013surveymultiviewmachine}
Shiliang Sun.
\newblock A survey of multi-view machine learning.
\newblock \emph{Neural Computing and Applications}, 23\penalty0 (7):\penalty0
  2031--2038, 2013.
\newblock ISSN 1433-3058.
\newblock \doi{10.1007/s00521-013-1362-6}.

\bibitem[Tseng et~al.(2021)Tseng, Zvonkov, Nakalembe, and
  Kerner]{tseng2021cropharvestglobaldataset}
Gabriel Tseng, Ivan Zvonkov, Catherine~Lilian Nakalembe, and Hannah Kerner.
\newblock {{CropHarvest}}: {{A}} global dataset for crop-type classification.
\newblock \emph{Proceedings of Conference on Neural Information Processing
  Systems (NIPS) Datasets and Benchmarks Track (Round 2)}, 2021.

\bibitem[Valente et~al.(2019)Valente, Doldersum, Roers, and
  Kooistra]{valente2019detectingrumexobtusifolius}
J.~Valente, M.~Doldersum, C.~Roers, and L.~Kooistra.
\newblock {{DETECTING}} {{{\emph{RUMEX OBTUSIFOLIUS}}}} {{WEED PLANTS IN
  GRASSLANDS FROM UAV RGB IMAGERY USING DEEP LEARNING}}.
\newblock In \emph{{{ISPRS Annals}} of the {{Photogrammetry}}, {{Remote
  Sensing}} and {{Spatial Information Sciences}}}, volume IV-2-W5, pages
  179--185. {Copernicus GmbH}, 2019.
\newblock \doi{10.5194/isprs-annals-IV-2-W5-179-2019}.

\bibitem[Valero et~al.(2019)Valero, Arnaud, Planells, Ceschia, and
  Dedieu]{valero2019sentinelclassifierfusion}
S.~Valero, L.~Arnaud, M.~Planells, E.~Ceschia, and G.~Dedieu.
\newblock Sentinel’s {{Classifier Fusion System}} for {{Seasonal Crop
  Mapping}}.
\newblock In \emph{{{IEEE International Geoscience}} and {{Remote Sensing
  Symposium}} (IGARSS)}, pages 6243--6246, 2019.
\newblock \doi{10.1109/IGARSS.2019.8898011}.

\bibitem[Vielzeuf et~al.(2018)Vielzeuf, Lechervy, Pateux, and
  Jurie]{vielzeuf2018centralnetmultilayerapproach}
Valentin Vielzeuf, Alexis Lechervy, Stephane Pateux, and Frederic Jurie.
\newblock {{CentralNet}}: A {{Multilayer Approach}} for {{Multimodal Fusion}}.
\newblock In \emph{Proceedings of the European Conference on Computer Vision
  (ECCV) Workshops}, pages 0--0, 2018.

\bibitem[Wang et~al.(2018)Wang, Tran, Desai, Lobell, and
  Ermon]{wang2018deeptransferlearninga}
Anna~X. Wang, Caelin Tran, Nikhil Desai, David Lobell, and Stefano Ermon.
\newblock Deep {{Transfer Learning}} for {{Crop Yield Prediction}} with
  {{Remote Sensing Data}}.
\newblock In \emph{Proceedings of the 1st {{ACM SIGCAS Conference}} on
  {{Computing}} and {{Sustainable Societies}}}, {{COMPASS}} '18, pages 1--5.
  {Association for Computing Machinery}, 2018.
\newblock ISBN 978-1-4503-5816-3.
\newblock \doi{10.1145/3209811.3212707}.

\bibitem[Wang et~al.(2021)Wang, Liu, Pei, Huang, Zhang, and
  Yang]{wang2021multiviewattentioncnnlstm}
Chenwei Wang, Xiaoyu Liu, Jifang Pei, Yulin Huang, Yin Zhang, and Jianyu Yang.
\newblock Multiview {{Attention CNN-LSTM Network}} for {{SAR Automatic Target
  Recognition}}.
\newblock \emph{IEEE Journal of Selected Topics in Applied Earth Observations
  and Remote Sensing}, 14:\penalty0 12504--12513, 2021.
\newblock ISSN 2151-1535.
\newblock \doi{10.1109/JSTARS.2021.3130582}.

\bibitem[Wang et~al.(2019)Wang, Boudreau, Luo, Tan, and
  Zhou]{wang2019deepmultiviewinformationa}
Qi~Wang, Claire Boudreau, Qixing Luo, Pang-Ning Tan, and Jiayu Zhou.
\newblock Deep {{Multi-view Information Bottleneck}}.
\newblock In \emph{Proceedings of the 2019 {{SIAM International Conference}} on
  {{Data Mining}} ({{SDM}})}, Proceedings, pages 37--45. {Society for
  Industrial and Applied Mathematics}, 2019.
\newblock \doi{10.1137/1.9781611975673.5}.

\bibitem[Wang et~al.(2020{\natexlab{a}})Wang, Tran, and
  Feiszli]{wang2020whatmakestraininga}
Weiyao Wang, Du~Tran, and Matt Feiszli.
\newblock What {{Makes Training Multi-Modal Classification Networks Hard}}?
\newblock In \emph{Proceedings of IEEE/CVF Conference on Computer Vision and
  Pattern Recognition (CVPR)}, pages 12695--12705, 2020{\natexlab{a}}.

\bibitem[Wang et~al.(2020{\natexlab{b}})Wang, Huang, Feng, and
  Yin]{wang2020winterwheatyieldb}
Xinlei Wang, Jianxi Huang, Quanlong Feng, and Dongqin Yin.
\newblock Winter {{Wheat Yield Prediction}} at {{County Level}} and
  {{Uncertainty Analysis}} in {{Main Wheat-Producing Regions}} of {{China}}
  with {{Deep Learning Approaches}}.
\newblock \emph{Remote Sensing}, 12\penalty0 (11):\penalty0 1744,
  2020{\natexlab{b}}.
\newblock ISSN 2072-4292.
\newblock \doi{10.3390/rs12111744}.

\bibitem[Wang et~al.(2020{\natexlab{c}})Wang, Huang, Sun, Xu, Rong, and
  Huang]{wang2020deepmultimodalfusion}
Yikai Wang, Wenbing Huang, Fuchun Sun, Tingyang Xu, Yu~Rong, and Junzhou Huang.
\newblock Deep {{Multimodal Fusion}} by {{Channel Exchanging}}.
\newblock \emph{Advances in Neural Information Processing Systems (NIPS)},
  33:\penalty0 4835--4845, 2020{\natexlab{c}}.

\bibitem[Waske and Benediktsson(2007)]{waske2007fusionsupportvector}
BjÖrn Waske and JÓn~Atli Benediktsson.
\newblock Fusion of {{Support Vector Machines}} for {{Classification}} of
  {{Multisensor Data}}.
\newblock \emph{IEEE Transactions on Geoscience and Remote Sensing},
  45\penalty0 (12):\penalty0 3858--3866, 2007.
\newblock ISSN 1558-0644.
\newblock \doi{10.1109/TGRS.2007.898446}.

\bibitem[Wu et~al.(2021)Wu, Hong, and
  Chanussot]{wu2021convolutionalneuralnetworks}
Xin Wu, Danfeng Hong, and Jocelyn Chanussot.
\newblock Convolutional {{Neural Networks}} for {{Multimodal Remote Sensing
  Data Classification}}.
\newblock \emph{IEEE Transactions on Geoscience and Remote Sensing}, pages
  1--1, 2021.
\newblock ISSN 1558-0644.
\newblock \doi{10.1109/TGRS.2021.3124913}.

\bibitem[Xie et~al.(2021)Xie, Han, Xie, Chen, and
  Chen]{xie2021multiviewfusionnetwork}
Lihong Xie, Ruiling Han, Songhong Xie, Dongjing Chen, and Yaxuan Chen.
\newblock Multi-{{View Fusion Network}} for {{Crop Disease Recognition}}.
\newblock In \emph{{{International Conference}} on {{Algorithms}},
  {{Computing}} and {{Systems}}}, {{ICACS}} '21, pages 121--126. {Association
  for Computing Machinery}, 2021.
\newblock ISBN 978-1-4503-8508-4.
\newblock \doi{10.1145/3490700.3490724}.

\bibitem[Xu et~al.(2018)Xu, Li, Ran, Du, Gao, and
  Zhang]{xu2018multisourceremotesensing}
Xiaodong Xu, Wei Li, Qiong Ran, Qian Du, Lianru Gao, and Bing Zhang.
\newblock Multisource {{Remote Sensing Data Classification Based}} on
  {{Convolutional Neural Network}}.
\newblock \emph{IEEE Transactions on Geoscience and Remote Sensing},
  56\penalty0 (2):\penalty0 937--949, 2018.
\newblock ISSN 1558-0644.
\newblock \doi{10.1109/TGRS.2017.2756851}.

\bibitem[Yalcin(2018)]{yalcin2018phenologyrecognitionusing}
Hulya Yalcin.
\newblock Phenology recognition using deep learning.
\newblock In \emph{{{Electric Electronics}}, {{Computer Science}}, {{Biomedical
  Engineerings}}' {{Meeting}} ({{EBBT}})}, pages 1--5, 2018.
\newblock \doi{10.1109/EBBT.2018.8391423}.

\bibitem[Yan et~al.(2021)Yan, Hu, Mao, Ye, and
  Yu]{yan2021deepmultiviewlearning}
Xiaoqiang Yan, Shizhe Hu, Yiqiao Mao, Yangdong Ye, and Hui Yu.
\newblock Deep multi-view learning methods: {{A}} review.
\newblock \emph{Neurocomputing}, 448:\penalty0 106--129, 2021.
\newblock ISSN 0925-2312.
\newblock \doi{10.1016/j.neucom.2021.03.090}.

\bibitem[Yang et~al.(2019)Yang, Shi, Han, Zha, and
  Zhu]{yang2019deepconvolutionalneurala}
Qi~Yang, Liangsheng Shi, Jinye Han, Yuanyuan Zha, and Penghui Zhu.
\newblock Deep convolutional neural networks for rice grain yield estimation at
  the ripening stage using {{UAV-based}} remotely sensed images.
\newblock \emph{Field Crops Research}, 235:\penalty0 142--153, 2019.
\newblock ISSN 0378-4290.
\newblock \doi{10.1016/j.fcr.2019.02.022}.

\bibitem[Yang et~al.(2017)Yang, Ramesh, Chitta, Madhvanath, Bernal, and
  Luo]{yang2017deepmultimodalrepresentation}
Xitong Yang, Palghat Ramesh, Radha Chitta, Sriganesh Madhvanath, Edgar~A.
  Bernal, and Jiebo Luo.
\newblock Deep {{Multimodal Representation Learning}} from {{Temporal Data}}.
\newblock In \emph{Proceedings of {{IEEE Conference}} on {{Computer Vision}}
  and {{Pattern Recognition}} ({{CVPR}})}, pages 5066--5074, 2017.
\newblock \doi{10.1109/CVPR.2017.538}.

\bibitem[You et~al.(2017)You, Li, Low, Lobell, and
  Ermon]{you2017deepgaussianprocessb}
Jiaxuan You, Xiaocheng Li, Melvin Low, David Lobell, and Stefano Ermon.
\newblock Deep {{Gaussian}} process for crop yield prediction based on remote
  sensing data.
\newblock In \emph{Proceedings of {{Conference}} on {{Artificial Intelligence}}
  (AAI)}, {{AAAI}}'17, pages 4559--4565. {AAAI Press}, 2017.

\bibitem[Yuan et~al.(2021)Yuan, Zhuang, Schaefer, Feng, Guan, and
  Fang]{yuan2021deeplearningbasedmultispectralsatellite}
Kunhao Yuan, Xu~Zhuang, Gerald Schaefer, Jianxin Feng, Lin Guan, and Hui Fang.
\newblock Deep-{{Learning-Based Multispectral Satellite Image Segmentation}}
  for {{Water Body Detection}}.
\newblock \emph{IEEE Journal of Selected Topics in Applied Earth Observations
  and Remote Sensing}, 14:\penalty0 7422--7434, 2021.
\newblock ISSN 2151-1535.
\newblock \doi{10.1109/JSTARS.2021.3098678}.

\bibitem[Yuan et~al.(2020)Yuan, Shen, Li, Li, Li, Jiang, Xu, Tan, Yang, Wang,
  Gao, and Zhang]{yuan2020deeplearningenvironmentala}
Qiangqiang Yuan, Huanfeng Shen, Tongwen Li, Zhiwei Li, Shuwen Li, Yun Jiang,
  Hongzhang Xu, Weiwei Tan, Qianqian Yang, Jiwen Wang, Jianhao Gao, and
  Liangpei Zhang.
\newblock Deep learning in environmental remote sensing: {{Achievements}} and
  challenges.
\newblock \emph{Remote Sensing of Environment}, 241:\penalty0 111716, 2020.
\newblock ISSN 0034-4257.
\newblock \doi{10.1016/j.rse.2020.111716}.

\bibitem[Zbontar et~al.(2021)Zbontar, Jing, Misra, LeCun, and
  Deny]{zbontar2021barlowtwinsselfsupervised}
Jure Zbontar, Li~Jing, Ishan Misra, Yann LeCun, and Stephane Deny.
\newblock Barlow {{Twins}}: {{Self-Supervised Learning}} via {{Redundancy
  Reduction}}.
\newblock In \emph{Proceedings of {{International Conference}} on {{Machine
  Learning}} (ICML)}, pages 12310--12320. {PMLR}, 2021.

\bibitem[Zhang et~al.(2020{\natexlab{a}})Zhang, Zhang, Luo, Cao, and
  Tao]{zhang2020combiningopticalfluorescencea}
Liangliang Zhang, Zhao Zhang, Yuchuan Luo, Juan Cao, and Fulu Tao.
\newblock Combining {{Optical}}, {{Fluorescence}}, {{Thermal Satellite}}, and
  {{Environmental Data}} to {{Predict County-Level Maize Yield}} in {{China
  Using Machine Learning Approaches}}.
\newblock \emph{Remote Sensing}, 12\penalty0 (1):\penalty0 21,
  2020{\natexlab{a}}.
\newblock ISSN 2072-4292.
\newblock \doi{10.3390/rs12010021}.

\bibitem[Zhang et~al.(2022)Zhang, Li, Tao, Li, and
  Du]{zhang2022informationfusionclassification}
Mengmeng Zhang, Wei Li, Ran Tao, Hengchao Li, and Qian Du.
\newblock Information {{Fusion}} for {{Classification}} of {{Hyperspectral}}
  and {{LiDAR Data Using IP-CNN}}.
\newblock \emph{IEEE Transactions on Geoscience and Remote Sensing},
  60:\penalty0 1--12, 2022.
\newblock ISSN 1558-0644.
\newblock \doi{10.1109/TGRS.2021.3093334}.

\bibitem[Zhang et~al.(2020{\natexlab{b}})Zhang, Du, Lin, Wang, Li, Xue, and
  Bai]{zhang2020hybridattentionawarefusion}
Peng Zhang, Peijun Du, Cong Lin, Xin Wang, Erzhu Li, Zhaohui Xue, and Xuyu Bai.
\newblock A {{Hybrid Attention-Aware Fusion Network}} ({{HAFNet}}) for
  {{Building Extraction}} from {{High-Resolution Imagery}} and {{LiDAR Data}}.
\newblock \emph{Remote Sensing}, 12\penalty0 (22):\penalty0 3764,
  2020{\natexlab{b}}.
\newblock ISSN 2072-4292.
\newblock \doi{10.3390/rs12223764}.

\bibitem[Zhao et~al.(2022)Zhao, Zhang, Shi, Zhou, Chen, Yao, and
  Xue]{zhao2022multisourcecollaborativeenhanced}
Jiaqi Zhao, Di~Zhang, Boyu Shi, Yong Zhou, Jingyang Chen, Rui Yao, and Yong
  Xue.
\newblock Multi-source collaborative enhanced for remote sensing images
  semantic segmentation.
\newblock \emph{Neurocomputing}, 493:\penalty0 76--90, 2022.
\newblock ISSN 0925-2312.
\newblock \doi{10.1016/j.neucom.2022.04.045}.

\bibitem[Zhao et~al.(2017)Zhao, Xie, Xu, and
  Sun]{zhao2017multiviewlearningoverview}
Jing Zhao, Xijiong Xie, Xin Xu, and Shiliang Sun.
\newblock Multi-view learning overview: {{Recent}} progress and new challenges.
\newblock \emph{Information Fusion}, 38:\penalty0 43--54, 2017.
\newblock ISSN 1566-2535.
\newblock \doi{10.1016/j.inffus.2017.02.007}.

\bibitem[Zhao et~al.(2020)Zhao, Tao, Li, Li, Du, Liao, and
  Philips]{zhao2020jointclassificationhyperspectral}
Xudong Zhao, Ran Tao, Wei Li, Heng-Chao Li, Qian Du, Wenzhi Liao, and Wilfried
  Philips.
\newblock Joint {{Classification}} of {{Hyperspectral}} and {{LiDAR Data Using
  Hierarchical Random Walk}} and {{Deep CNN Architecture}}.
\newblock \emph{IEEE Transactions on Geoscience and Remote Sensing},
  58\penalty0 (10):\penalty0 7355--7370, 2020.
\newblock ISSN 1558-0644.
\newblock \doi{10.1109/TGRS.2020.2982064}.

\bibitem[Zheng et~al.(2022)Zheng, Wu, Huan, He, and
  Zhang]{zheng2022gathertoguidenetworkremote}
Xianwei Zheng, Xiujie Wu, Linxi Huan, Wei He, and Hongyan Zhang.
\newblock A {{Gather-to-Guide Network}} for {{Remote Sensing Semantic
  Segmentation}} of {{RGB}} and {{Auxiliary Image}}.
\newblock \emph{IEEE Transactions on Geoscience and Remote Sensing},
  60:\penalty0 1--15, 2022.
\newblock ISSN 1558-0644.
\newblock \doi{10.1109/TGRS.2021.3103517}.

\bibitem[Zhou et~al.(2022)Zhou, Jin, Lei, and
  Hwang]{zhou2022cegfnetcommonextraction}
Wujie Zhou, Jianhui Jin, Jingsheng Lei, and Jenq-Neng Hwang.
\newblock {{CEGFNet}}: {{Common Extraction}} and {{Gate Fusion Network}} for
  {{Scene Parsing}} of {{Remote Sensing Images}}.
\newblock \emph{IEEE Transactions on Geoscience and Remote Sensing},
  60:\penalty0 1--10, 2022.
\newblock ISSN 1558-0644.
\newblock \doi{10.1109/TGRS.2021.3109626}.

\bibitem[Zhu et~al.(2018)Zhu, Cai, Tian, and
  Williams]{zhu2018spatiotemporalfusionmultisourcea}
Xiaolin Zhu, Fangyi Cai, Jiaqi Tian, and Trecia Kay-Ann Williams.
\newblock Spatiotemporal {{Fusion}} of {{Multisource Remote Sensing Data}}:
  {{Literature Survey}}, {{Taxonomy}}, {{Principles}}, {{Applications}}, and
  {{Future Directions}}.
\newblock \emph{Remote Sensing}, 10\penalty0 (4):\penalty0 527, 2018.
\newblock ISSN 2072-4292.
\newblock \doi{10.3390/rs10040527}.

\end{thebibliography}

\clearpage
\setcounter{table}{0}
\renewcommand{\thetable}{A\arabic{table}}
\appendix

\section{Additional Tables} \label{sec:appendix}
Table~\ref{sup:tab:where_fuse} contains the data used to generate the Figure~\ref{fig:where_fuse} on the main content of this manuscript. While Table~\ref{sup:merge_func} contains the references from which the Table~\ref{tab:merge_func} on the main content of this manuscript was obtained.  Last, Table~\ref{sup:tab:fusion_strategies} contains the references of papers of the categorization made in Section~\ref{sec:questions}.

\begin{table}[!ht] 
\caption{Data used to generate Figure~\ref{fig:where_fuse}. Scenario correspond to the type of views used in each configuration where the results were obtained. The numbers correspond to: 1: best, 2: in-between, 3: worst, of the predictive performance when comparing the fusion strategies (input, feature, and decision level fusion). In parentheses different data sets or scenarios from the same document are indicated.}
\label{sup:tab:where_fuse}
\newcolumntype{C}{>{\centering\arraybackslash}X}
\footnotesize
\begin{tabularx}{\textwidth}{lCCC}
\hline
\textbf{Views scenario} (Reference) &  \textbf{Input}	& \textbf{Feature}  & \textbf{Decision} \\
\hline
optical-RGB, optical (NIR) (\cite{sa2016deepfruitsfruitdetection}) & 2 & - & 1 \\
optical-MS, DSM (\cite{chen2017deepfusionremotea}, San Francisco) & 2 & 1 & - \\
optical-HS, LiDAR (\cite{chen2017deepfusionremotea}, Houston) & 2 & 1 & - \\
optical-MS, DSM (\cite{audebert2018rgbveryhigh}) & - & 1 & 2 \\
optical-HS, LiDAR (\cite{xu2018multisourceremotesensing}, Houston, Trento) & 2 & 1 & - \\
optical-HS, optical-RGB (\cite{xu2018multisourceremotesensing}, Pavia, Salinas) & 2 & 1 & - \\
multiple optical-RGB (\cite{srivastava2020finegrainedlandusecharacterization}) & - & 1 & 2 \\
soil, cultivation, yield stats (\cite{livieris2020multipleinputneuralnetworka}) & 2 & 1 & - \\
optical-RGB, DSM (\cite{zhang2020hybridattentionawarefusion}) & - & 1 & 2 \\
optical-MS, thermal, DSM (\cite{maimaitijiang2020soybeanyieldpredictiona}) & 2 & 1 & - \\
optical-HS, LiDAR (\cite{wu2021convolutionalneuralnetworks}) & 3 & 1 & 2 \\
optical-MS, SAR (\cite{ofori-ampofo2021croptypemappingb}) & 1 & 2 & 3 \\
optical, SAR (\cite{cuelarosa2021investigatingfusionstrategiesa}) & - & 2 & 1 \\
optical-HS, LiDAR (\cite{hong2021morediversemeansa}, Houston) & 3 & 1 & 2 \\
optical-MS, SAR (\cite{hong2021morediversemeansa}, LCZ) & 3 & 1 & 2 \\
optical-MS, SAR (\cite{sebastianelli2021paradigmselectiondata}) & 3 & 1 & 2 \\
optical-RGB, DSM (\cite{hosseinpour2022cmgfnetdeepcrossmodala}) & - & 1 & 2 \\
optical-MS, SAR (\cite{saintefaregarnot2022multimodaltemporalattention}, Classification) & 2 & 1 & 3 \\
optical-MS, SAR (\cite{saintefaregarnot2022multimodaltemporalattention}, Segmentation)  & 2 & 1 & 2 \\
optical-MS, SAR (\cite{saintefaregarnot2022multimodaltemporalattention}, Panoptic)  & 1 & 2 & - \\
\hline
\end{tabularx}
\end{table}
\unskip

\begin{table}[!ht] 
\caption{Complete references for the works indicated in Table~\ref{tab:merge_func}.}\label{sup:merge_func}
\newcolumntype{C}{>{\centering\arraybackslash}X}
\newcolumntype{L}{>{\raggedright\arraybackslash}X}
\newcolumntype{R}{>{\raggedleft\arraybackslash}X}
\footnotesize
\begin{tabularx}{\textwidth}{lL}
\hline
\textbf{Name}	&  \textbf{Reference} \\
\hline
Concatenation & \cite{marmanis2016semanticsegmentationaerial,chen2017deepfusionremotea,xu2018multisourceremotesensing,pei2018sarautomatictarget,audebert2018rgbveryhigh,benedetti2018textfusiondeeplearninga,dealwis2019duoattentiondeep,gangopadhyay2019deeptimeseries,nguyen2019spatialtemporalmultitasklearninga,yang2019deepconvolutionalneurala,srivastava2019understandingurbanlanduse,rudner2019multi3netsegmentingflooded,ienco2019combiningsentinel1sentinel2b,khaki2020cnnrnnframeworkcropa,wang2020winterwheatyieldb,zhang2020combiningopticalfluorescencea,maimaitijiang2020soybeanyieldpredictiona,mohla2020fusatnetdualattention,hang2020classificationhyperspectrallidar,chu2020endtoendmodelricea,gadiraju2020multimodaldeeplearninga,livieris2020multipleinputneuralnetworka,irvin2020forestnetclassifyingdrivers,wu2021convolutionalneuralnetworks,yuan2021deeplearningbasedmultispectralsatellite,ofori-ampofo2021croptypemappingb,cao2021integratingmultisourcedata,hong2021morediversemeansa,sebastianelli2021paradigmselectiondata,dimartino2021multibranchdeeplearning,hong2021multimodalremotesensing,shahhosseini2021cornyieldpredictiona,li2022outcome2021ieee,zhang2022informationfusionclassification,saintefaregarnot2022multimodaltemporalattention,zhao2022multisourcecollaborativeenhanced} \\
Attention & \cite{mohla2020fusatnetdualattention,zheng2022gathertoguidenetworkremote,zhou2022cegfnetcommonextraction,gao2022adaptivespectralspatialfeature,zhao2022multisourcecollaborativeenhanced} \\
Uniform-sum	& \cite{luus2015multiviewdeeplearning,audebert2017semanticsegmentationeartha,ahmad2017cnnganbased,audebert2018rgbveryhigh,mao2018deepcrossmodalretrieval,srivastava2019understandingurbanlanduse,liu2020multiviewselfconstructinggrapha,zhang2020hybridattentionawarefusion,srivastava2020finegrainedlandusecharacterization,hang2020classificationhyperspectrallidar,gadiraju2020multimodaldeeplearninga,wu2021convolutionalneuralnetworks,ofori-ampofo2021croptypemappingb,sebastianelli2021paradigmselectiondata,cuelarosa2021investigatingfusionstrategiesa,saintefaregarnot2022multimodaltemporalattention} \\
Weighted-sum & \cite{hang2020classificationhyperspectrallidar,rashkovetsky2021wildfiredetectionmultisensor,robinson2021globallandcovermapping,sebastianelli2021paradigmselectiondata,cuelarosa2021investigatingfusionstrategiesa,zheng2022gathertoguidenetworkremote,zhao2022multisourcecollaborativeenhanced} \\ 
Gated & \cite{zhang2020hybridattentionawarefusion,cao2021c3netcrossmodalfeaturea,hosseinpour2022cmgfnetdeepcrossmodala} \\
Product	&  \cite{valero2019sentinelclassifierfusion,ofori-ampofo2021croptypemappingb} \\
Maximum &  \cite{srivastava2020finegrainedlandusecharacterization,hang2020classificationhyperspectrallidar} \\
Majority & \cite{waske2007fusionsupportvector,liu2018deepconvolutionalneural,zhang2020hybridattentionawarefusion} \\
\hline
\end{tabularx}
\end{table}
\unskip

\begin{table}[!ht] 
\caption{Categorization of papers reviewed, the categorization is non-exclusive. The discussion about each type of categorization could be found in Section~\ref{sec:questions}.}\label{sup:tab:fusion_strategies}
\newcolumntype{L}{>{\raggedright\arraybackslash}X}
\footnotesize
\begin{tabularx}{\textwidth}{lL}
\hline
\textbf{Categorization}	& \textbf{References} \\
\hline
Input-level fusion & \cite{benediktsson1989neuralnetworkapproaches,kumar2002estimatingevapotranspirationusing,gomez-chova2006urbanmonitoringusing,nemmour2006multiplesupportvector,kumar2008comparativestudyconventional,camps-valls2008kernelbasedframeworkmultitemporal,russ2008dataminingneural,debes2014hyperspectrallidardata,chen2014deeplearningbasedclassification,johnson2014assessmentprewithinseasona,dona2015integratedsatellitedata,khodadadzadeh2015fusionhyperspectrallidar,ahamed2015applyingdatamining,bocca2016effecttuningfeaturea,kim2016machinelearningapproaches,ghamisi2017hyperspectrallidardata,albughdadi2017missingdatareconstruction,nijhawan2017deeplearninghybrid,kussul2017deeplearningclassificationa,you2017deepgaussianprocessb,h.russello2018convolutionalneuralnetworksa,wang2018deeptransferlearninga,jiang2018predictingcountylevel,denize2019evaluationusingsentinel1,gomez2019potatoyieldpredictionb,sun2019countylevelsoybeanyielda,niedbala2019simplemodelbaseda,cai2019integratingsatelliteclimatea,rambour2020flooddetectiontime,pageot2020detectionirrigatedrainfed,maimaitijiang2020soybeanyieldpredictiona,lin2020deepcropnetdeepspatialtemporal,peng2020assessingbenefitsatellitebaseda,kang2020comparativeassessmentenvironmentala,bhojani2020wheatcropyield,jiang2020deeplearningapproacha,konapala2021exploringsentinel1sentinel2,sebastianelli2021paradigmselectiondata,robinson2021globallandcovermapping,song2021evaluationlandsatsentinel2,sagan2021fieldscalecropyielda,ofori-ampofo2021croptypemappingb,gavahi2021deepyieldcombinedconvolutionala,meng2021predictingmaizeyielda,khaki2021simultaneouscornsoybeana,hong2021morediversemeansa,diaconu2022understandingroleweather,mantsis2022multimodalfusionsentinel,saintefaregarnot2022multimodaltemporalattention} \\ \hline
Feature-level fusion & sub-feature based:  \cite{marmanis2016semanticsegmentationaerial,rudner2019multi3netsegmentingflooded,mohla2020fusatnetdualattention,ofori-ampofo2021croptypemappingb,wang2021multiviewattentioncnnlstm,hong2021morediversemeansa,zheng2022gathertoguidenetworkremote,zhang2022informationfusionclassification,saintefaregarnot2022multimodaltemporalattention}, embedding based: \cite{chen2017deepfusionremotea,mao2018deepcrossmodalretrieval,xu2018multisourceremotesensing,benedetti2018textfusiondeeplearninga,srivastava2019understandingurbanlanduse,ienco2019combiningsentinel1sentinel2b,yang2019deepconvolutionalneurala,nguyen2019spatialtemporalmultitasklearninga,dealwis2019duoattentiondeep,gangopadhyay2019deeptimeseries,irvin2020forestnetclassifyingdrivers,srivastava2020finegrainedlandusecharacterization,khaki2020cnnrnnframeworkcropa,maimaitijiang2020soybeanyieldpredictiona,zhang2020combiningopticalfluorescencea,chu2020endtoendmodelricea,livieris2020multipleinputneuralnetworka,gadiraju2020multimodaldeeplearninga,wang2020winterwheatyieldb,yuan2021deeplearningbasedmultispectralsatellite,dimartino2021multibranchdeeplearning,hong2021multimodalremotesensing,wu2021convolutionalneuralnetworks,ofori-ampofo2021croptypemappingb,shahhosseini2021cornyieldpredictiona,xie2021multiviewfusionnetwork,hong2021morediversemeansa,gao2022adaptivespectralspatialfeature} \\ \hline  
Decision-level fusion & \cite{audebert2017semanticsegmentationeartha,audebert2018rgbveryhigh,liu2020multiviewselfconstructinggrapha,sebastianelli2021paradigmselectiondata,wu2021convolutionalneuralnetworks,ofori-ampofo2021croptypemappingb,saintefaregarnot2022multimodaltemporalattention} \\ \hline \hline
Ensemble-based fusion & \cite{waske2007fusionsupportvector,luus2015multiviewdeeplearning,ahmad2017cnnganbased,liu2018deepconvolutionalneural,valero2019sentinelclassifierfusion,rashkovetsky2021wildfiredetectionmultisensor,ma2021outcome2021ieee,robinson2021globallandcovermapping,li2022outcome2021ieee} \\
Hybrid fusion & \cite{hang2020classificationhyperspectrallidar,zhang2020hybridattentionawarefusion,cao2021integratingmultisourcedata,cuelarosa2021investigatingfusionstrategiesa,li2022outcome2021ieee} \\
Dense fusion & \cite{pei2018sarautomatictarget,audebert2018rgbveryhigh,zhang2020hybridattentionawarefusion,cao2021c3netcrossmodalfeaturea,zhou2022cegfnetcommonextraction,zhao2022multisourcecollaborativeenhanced,hosseinpour2022cmgfnetdeepcrossmodala} \\
\hline
\end{tabularx}
\end{table}
\unskip

\end{document}